\documentclass[letterpaper, 10 pt, conference]{ieeeconf}  % Comment this line out if you need a4paper

\usepackage[compatibility=false]{caption}
\IEEEoverridecommandlockouts
\usepackage[space, compress, sort]{cite}

\usepackage{graphics} % for pdf, bitmapped graphics files
\usepackage{epsfig} % for postscript graphics files
\usepackage{mathptmx} % assumes new font selection scheme installed
\usepackage{times} % assumes new font selection scheme installed
\usepackage{amsmath} % assumes amsmath package installed
\usepackage{todonotes}

\usepackage{amssymb, mathtools}  % assumes amsmath package installed
\usepackage{float} % allows [H] float placement
\makeatletter
\def\@opargbegintheorem#1#2#3{\trivlist
   \item[]{\bfseries #1\ #2\ (#3)} \itshape}
\makeatother

\usepackage{float}
\usepackage[breaklinks=true,bookmarks=true,colorlinks]{hyperref}
\usepackage{graphicx}
\usepackage{booktabs}   % for \toprule, \midrule, \
\usepackage{multirow}   % for \multirow cells
\usepackage{xcolor}
\usepackage{multirow}  % For merging rows
\usepackage{adjustbox}
\usepackage{float}
\usepackage[font=small]{caption}
\usepackage{algorithm}
\usepackage{algorithmic}
\usepackage{xspace}
\usepackage{dsfont} % for \mathds

\newcommand{\added}[1]{{\color{black}{#1}}}
\newcommand{\addedv}[1]{{\color{black}{#1}}}
\newcommand{\addedar}[1]{{\color{black}{#1}}}

\newcommand{\method}{CLAE\xspace}

\hypersetup{hidelinks}

\title{\LARGE \bf Steering Multirobot Behavior via Closed-Loop Affine Activation Editing}

\author{
Satyajeet Das$^{1}$, Darren Chiu$^{1}$, Shashank Hegde$^{2}$,
and Gaurav S. Sukhatme$^{1}$%
\thanks{$^{1}$Satyajeet Das, Darren Chiu, and Gaurav S. Sukhatme
are with the Department of Computer Science,
University of Southern California, Los Angeles, CA 90089, USA.
Emails: {\tt\small \{satyajee, chiudarr, gaurav\}@usc.edu}.}%
\thanks{$^{2}$Shashank Hegde is with NVIDIA, CA, USA.
Email: {\tt\small shashankh@nvidia.com}.}%
}

\begin{document}

\maketitle
\thispagestyle{empty}
\pagestyle{empty}

%%%%%%%%%%%%%%%%%%%%%%%%%%%%%%%%%%%%%%%%%%%%%%%%%%%%%%%%%%%%%%%%%%%%%%%%%%%%%%%%

\begin{abstract}
Real-world robots need to adapt their behavior beyond the envelope of their pre-trained policy. Policy finetuning or retraining are options, but they risk catastrophic forgetting, degrading the pretrained policy's base performance. 
To combat this, we introduce  \textbf{CLAE}: \emph{\underline{C}losed-\underline{L}oop Affine \underline{A}ctivation \underline{E}diting}, an inference-time framework for steering the behavior of a frozen policy by editing intermediate activations while keeping the base policy weights and downstream action head untouched. CLAE approaches behavior steering as a closed-loop problem whose outputs edit policy activations that adapt online to the robot state, environment, target behavior, and multi-robot context. It trains a sparse autoencoder over frozen-policy activations, selects behavior-relevant latent features via post-hoc probing, and learns a lightweight RL-based steering policy that applies state-dependent affine edits to selected latents during inference. We validate CLAE on a frozen multi-quadrotor navigation policy trained to perform a single task: navigating robots to a set of goal locations while avoiding obstacles. Through extensive simulations and physical tests, we show that while navigating to their goal positions, CLAE can 1. steer individual robot behavior by controlling each robot's velocity profile; 2. coordinate multirobot behavior by preserving a desired formation, and 3. produce entirely new behavior wherein robots are required to reduce their exposure to surveillance cameras in the environment. 
% Our project website is available at \url{https://claerobotics.github.io/}.
Our project website is available at \href{https://clae-robotics.github.io/}{https://clae-robotics.github.io/}.
\end{abstract}

%===============================================================================
% \input{sections/Introduction}
\section{Introduction}
\label{sec:introduction}

% End-to-end learned robot policies are increasingly used to map high-dimensional % observations, task context, and robot state directly to actions. 
%Such policies
Learned robot policies are commonly obtained via imitation learning on robot demonstrations, reinforcement learning from environment interactions, or hybrid pipelines that combine large-scale robot data, pretrained vision-language models, and task-specific supervision. %\citep{o2024open,kim2024openvla,huang2023quadrotor}. 
When deployed, a trained policy expresses a set of behaviors (a "behavioral envelope") from its training data, objective, conditioning interface, and deployment assumptions. In practice, robots need to modify or extend their behavioral envelope when deployed: a mobile robot or manipulator trained for aggressive navigation or goal reaching may need to behave more cautiously around people, a legged robot trained for energy-efficient navigation may need to trade efficiency for speed, and a team of flying robots trained to navigate to a goal may need to track a new velocity profile, maintain a formation, or avoid certain regions. 

%Robot deployment requires expanding the envelope of existing behaviors, whether encoded as purely learned policies or otherwise. 
Fine-tuning or retraining may adapt a policy to such requirements, but it changes the base policy weights, risking catastrophic forgetting
% requires additional demonstrations/simulated interaction, 
and necessitates revalidation of the %updated 
policy. 
% Weight updates may also alter previously learned behaviors. 
These concerns become more pronounced as the community moves towards generalist pretrained models, 
% where maintaining separate fine-tuned variants for every behavioral preference is increasingly unattractive.
\added{where maintaining separate fine-tuned variants for every behavioral preference adds deployment and revalidation cost.}
The multirobot setting amplifies this challenge: the correct adaptation for each robot depends not only on its own state but also on teammates' states. These considerations motivate our central question: can we modify/expand the inference-time behavioral envelope of a trained robot policy while keeping the base policy fixed?

\begin{figure*}[t]
    \centering
    \includegraphics[width=1\linewidth]
    {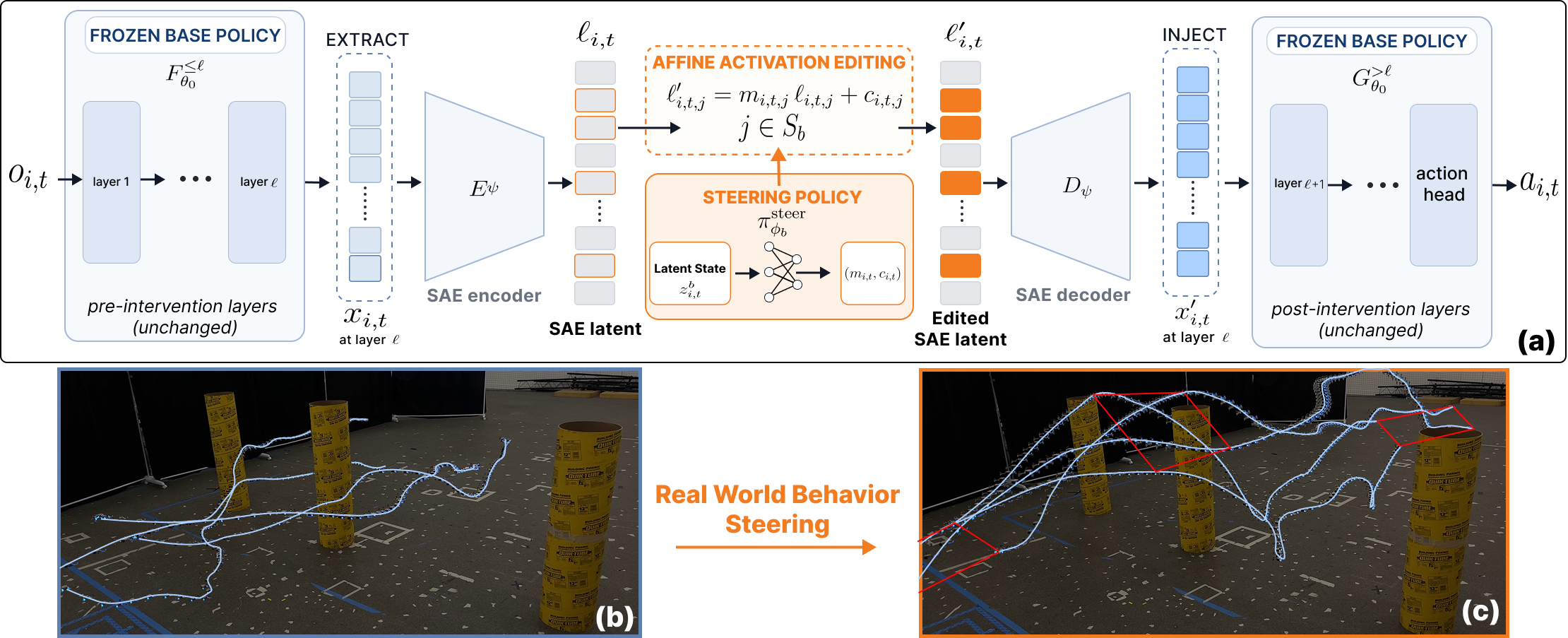}

    % \caption{\textbf{Closed-Loop Affine Activation Editing (CLAE).}
    % An intermediate activation $x_{i,t}$ from the frozen base policy is encoded into a sparse latent $\ell_{i,t}$, edited by a state-dependent affine map on the behavior-relevant latent$S_b$, and decoded back into the frozen policy. A lightweight PPO-trained steering policy $\pi^{\mathrm{steer}}_{\phi_b}$ produces the affine edit $\xi_{i,t} = (m_{i,t}, c_{i,t})$ online. Frozen components in blue, steering policy in green, selected latents in orange (outlines: pre-edit; filled: post-edit).} 
      \caption{\textbf{Closed-Loop Affine Activation Editing (CLAE).}
        \textbf{(a)} An intermediate activation $x_{i,t}$ from the frozen base policy is encoded into SAE latents $\ell_{i,t}$. The steering policy $\pi^{\mathrm{steer}}_{\phi_b}$ outputs $(m_{i,t},c_{i,t})$, which applies the affine edit
    $\ell'_{i,t,j}=m_{i,t,j}\ell_{i,t,j}+c_{i,t,j}$ on the select latent set $\mathcal{S}_b$, while unselected latents pass through unchanged. The edited latents are decoded and inserted back into the frozen policy. Bottom panels show real-world deployment of four quadrotors on the multi-robot formation task. In \textbf{(b}), the base policy navigates independently, dispersing around obstacles. In  \textbf{(c)}, CLAE edits activations to steer the quadrotors toward the formation geometry (red overlays), balancing the effort to stay in formation with the core collision-avoidance and goal-reaching behavior of the underlying policy.}
    \label{fig:clae_overview}
\end{figure*}

% We study this problem as \emph{post-training behavior steering}: modifying the closed-loop behavior of a trained policy at inference time while preserving its base competence. The key distinction is between task competence and deployment behavior. A policy may already know how to navigate, avoid obstacles, and reach goals, yet still need to express those skills under a different behavioral envelope. 
% Recent inference-time steering work makes a similar point for pretrained robot policies under spatial or semantic shifts: failures may reflect the absence of a mechanism for selectively adapting existing skills at test time, rather than the absence of the underlying motor skill itself \citep{liu2026vls}. We ask whether such adaptation can be achieved by editing the internal activations of a frozen robot policy.

% We study this problem as \emph{post-training behavior steering}: modifying the closed-loop behavior of a trained policy at inference time while preserving its base competence. 
\added{We study this problem as \emph{post-training behavior steering}: modifying the closed-loop behavior of a trained policy at inference time while aiming to retain its base competence.}
Recent inference-time steering work~\cite{liu2026vls} makes a similar point for pretrained robot policies under spatial or semantic shifts: failures may reflect the absence of a mechanism for selectively adapting existing skills at test time, rather than the absence of the underlying motor skill itself. Existing post-training steering methods~\cite{nakamoto2024vgps,wang2024itps,wagenmaker2025dsrl,liu2026vls,wu2025forewarn,chen2026steerable} typically act around the policy output, by modifying prompts, command interfaces, sampling trajectories, latent noise, candidate actions, or action selection. In contrast, activation steering modifies behavior by editing intermediate representations within the frozen policy. Prior work in language models~\cite{turner2023actadd,zou2023representation,cunningham2023sparse,templeton2024scaling} shows that internal activations can be modified to steer outputs. Recent robotics work~\cite{haon2025mechanistic,swann2026sparse,das2025latent} extends this idea to robot policies, demonstrating scalar, direction-based, or safety-specific interventions. However, these methods do not formulate activation editing as closed-loop behavior steering, in which edit parameters vary online with the robot's state, environment, target behavior, and other robots.

% However, existing robot activation-editing methods primarily demonstrate scalar,
% direction-based, or safety-specific steerability \citep{haon2025mechanistic,swann2026sparse,das2025latent}. They do not formulate
% activation editing as closed-loop behavior steering, where edit parameters must
% vary online as the robot state, environment, target behavior, and other agents
% evolve.

% We propose \emph{closed-loop affine activation editing} (\method), a framework for steering the behavior of a frozen robot policy by editing intermediate activations rather than updating weights or adding an action-residual controller. At each policy step, we encode an intermediate activation block into a sparse autoencoder (SAE) basis, apply affine edits to a small set of behavior-relevant SAE latents, and insert the decoded edited activation back into the frozen policy. A lightweight steering policy predicts the affine edit parameters online; the base policy's frozen downstream layers still produce the final robot action.

We propose \emph{closed-loop affine activation editing} (\method), a framework for steering a frozen robot policy by editing intermediate activations rather than updating weights or adding an action-residual controller. At each policy step, we encode an intermediate activation block into a sparse autoencoder (SAE) basis, apply affine edits to a small set of behavior-relevant SAE latents, and insert the decoded edited activation back into the frozen policy. A lightweight steering policy predicts the affine edit parameters online; the base policy's frozen downstream layers produce the 
% final 
robot action.

We evaluate \method on a frozen multi-quadrotor navigation policy across three behavior-steering tasks: velocity-profile tracking, multi-robot formation control, and camera-aware stealth navigation. Together, these test whether \method can steer robot-level motion, coordinate team behavior, and inject deployment-time behavior without making any modifications to the underlying policy. \method is the first framework for inference-time, closed-loop behavior steering of a multirobot policy through learned affine activation editing. 
\added{Although we validate the framework on a multirobot policy learned using reinforcement learning, \method only requires white-box (knowledge of architecture and weights) access to intermediate activations; therefore it is possible to steer arbitrary architectures, behaviors, and robot morphologies. Our hope is that \method can then be applied to behaviorally steer policies outside of our provided experiments.}

Our contributions are:
\begin{enumerate}
    \item We introduce \emph{closed-loop affine activation editing} (\method), a post-training behavior-steering framework that modifies intermediate activations at inference time while preserving the base policy weights.
    \item We build a sparse activation-editing interface by training an SAE on frozen-policy activations and using post-hoc behavior probes to select a compact set of behavior-relevant SAE latents. A lightweight RL steering policy then outputs state-dependent affine edits (multiplicative gains and additive offsets) over those latents at each policy step.
    % \item We build a sparse activation-editing interface by training an SAE on frozen-policy activations and using post-hoc behavior probes to select a compact set of behavior-relevant SAE latent features.
    % \item We train a lightweight RL steering policy to output state-dependent affine edits, consisting of multiplicative gains and additive offsets, for the selected SAE latent features at each policy step.
    \item We evaluate \method on a frozen multi-quadrotor navigation policy across three behaviors spanning robot-level motion, coordinated multi-robot behavior, and deployment-time behavior injection: velocity-profile tracking, multi-robot formation control, and stealth navigation, with validation in large-scale simulation and hardware experiments.
\end{enumerate}

\section{Related Work}
\label{sec:related_work}

\textbf{Post-training steering of frozen robot policies.}
A growing body of work studies deployment-time adaptation of frozen robot policies. Value-guided policy steering reranks candidate actions using an offline value function~\cite{nakamoto2024vgps}; inference-time policy steering biases generative-policy sampling with human interaction signals~\cite{wang2024itps}; and diffusion steering optimizes the latent-noise space of frozen diffusion policies with reinforcement learning~\cite{wagenmaker2025dsrl}. Recent methods also guide frozen diffusion or flow policies with VLM-derived rewards~\cite{liu2026vls}, select candidate plans using predicted outcomes and VLM reasoning~\cite{wu2025forewarn}, or improve controllability by training VLAs to follow richer command abstractions~\cite{chen2026steerable}. \added{A separate line of work leaves the policy weights untouched and instead learns a corrective term in action space~\cite{johannink2018residual,silver2018residual}. These methods demonstrate the importance of deployment-time steering, but primarily act through prompts, commands, sampling, latent noise, candidate actions, action selection, or action residuals. In contrast, we steer behavior by editing intermediate activations inside the frozen policy.}

% \textbf{Activation and representation steering.}
% Activation steering provides a complementary mechanism for modifying model behavior without weight updates. In language models, activation engineering modifies intermediate activations at inference time to steer model outputs \cite{turner2023actadd}, while representation engineering treats internal representations as objects that can be monitored and manipulated \cite{zou2023representation}. Sparse autoencoders provide sparse feature bases for analyzing and intervening on dense activations \cite{cunningham2023sparse,templeton2024scaling}, and representation surgery studies feature transformations beyond simple additive directions \cite{singh2024representation_surgery}. Our work brings these inference-time representation-intervention ideas to robot policies, where activation edits affect not only the current action but also the future observations encountered by the closed-loop system.

\textbf{Activation and representation steering.}
Activation steering provides a complementary mechanism for modifying model behavior without weight updates. In language models, activation engineering modifies intermediate activations at inference time~\cite{turner2023actadd}, while representation engineering studies internal representations as objects that can be monitored and manipulated~\cite{zou2023representation}. Sparse autoencoders provide sparse feature bases for analyzing and intervening on dense activations~\cite{cunningham2023sparse,templeton2024scaling}, and representation surgery studies feature transformations beyond additive directions~\cite{singh2024representation_surgery}. 
% Our work brings these representation-intervention ideas to robot policies, where activation edits affect not only the current action but also future observations induced by a closed-loop system.
% execution.
\added{Our work brings these representation-intervention ideas to robot policies, where activation edits affect not only the current action but also the future observations the edited policy induces.}

\textbf{Activation editing for robot behavior steering.}
Recent robotics work shows that internal activations of robot policies can be causally linked to behavior. Mechanistic interpretability methods identify semantic directions such as speed and direction in VLA models and steer them with activation interventions~\cite{haon2025mechanistic}; SAE-based VLA analyses show that sparse features can influence closed-loop robot behavior through decoder-direction interventions~\cite{swann2026sparse}; feature-controllability methods use linear probes and minimal raw-representation interventions to steer VLA outputs toward desired feature values~\cite{buurmeijer2026observing}; and Latent Activation Editing modifies activations of a frozen multirobot navigation policy to improve safety~\cite{das2025latent}. 
% These works demonstrate that robot behavior can be steered through internal activations. In contrast, \method learns a closed-loop affine edit policy whose parameters vary online with robot state, environment, neighboring agents, and target behavior, enabling behavior steering of a frozen multirobot policy without weight updates or action-residual commands. 
\added{While these methods enable inference-time activation intervention, in some
cases adaptively, none learns an activation-editing policy from
behavior-specific return. \method instead learns a closed-loop editing policy
that produces state-dependent affine edits conditioned on robot state,
environment, neighboring agents, and target behavior, while keeping the base
policy weights fixed.}

% \textbf{Activation editing for robot behavior steering.}
% Recent robotics work shows that internal activations of robot policies can be causally linked to behavior. Mechanistic interpretability methods for vision-language-action models identify semantic directions such as speed and direction and steer them with scalar activation interventions \cite{haon2025mechanistic}; SAE-based VLA analyses show that sparse features can causally influence closed-loop robot behavior through decoder-direction interventions \cite{swann2026sparse}; feature-controllability methods use linear probes and minimal raw-representation interventions to steer VLA outputs toward desired feature values \cite{buurmeijer2026observing}; and Latent Activation Editing modifies activations of a frozen multirobot navigation policy to improve safety \cite{das2025latent}. These works demonstrate that robot behavior can be steered by modifying internal activations rather than policy weights. In contrast, we learn a closed-loop affine edit policy whose parameters vary online with robot state, environment, neighboring agents, and target behavior, enabling fine-grained behavior control of a frozen multirobot policy without weight updates or action-residual commands.
%===============================================================================
% \input{sections/Method}
% \input{sections/Method1}
% \vspace{-6pt}
\section{Method}
\label{sec:method}
% \vspace{-0.5em}

We propose \emph{closed-loop affine activation editing} (\method), an inference-time mechanism for steering a frozen multirobot policy $\pi_{\theta_0}$(Figure~\ref{fig:clae_overview}a). At each policy step, we intercept an intermediate activation of the frozen base policy $\pi_{\theta_0}$ and encode it into a sparse autoencoder (SAE) basis, apply an affine edit to a small target-behavior-specific subset of SAE latents, decode the edited latent, and insert it back into the frozen policy. The base policy parameters $\theta_0$ remain fixed throughout: we do not fine-tune the policy, modify its architecture, or train an action-residual controller. For each target behavior $b$, rollout-derived metrics select a behavior-relevant SAE latent set $\mathcal{S}_b$, and a behavior-specific reward trains a lightweight steering policy $\pi_{\phi_b}^{\mathrm{steer}}$ that chooses affine edits online. The base policy, SAE, and steering policy are decentralized and shared across all robots.

\subsection{Frozen-Policy Activation Interface}
\label{sec:frozen_activation_interface}

For a given intervention layer $\ell$, we view the frozen base policy as a composition of the computation before $\ell$ ($F_{\theta_0}^{\leq \ell}$) and after $\ell$ ($G_{\theta_0}^{>\ell}$). Therefore, the base policy can be written as: $
    \pi_{\theta_0}
    =
    G_{\theta_0}^{>\ell}
    \circ
    F_{\theta_0}^{\leq \ell}.$
Given the robot observation $o_{i,t}$, the
frozen computation up to layer $\ell$ produces the intervention activation
$x_{i,t}$, and the remaining frozen layers map this activation to the robot action:
\begin{equation}
    x_{i,t}
    =
    F_{\theta_0}^{\leq \ell}(o_{i,t}),
    \qquad
    u_{i,t}
    =
    G_{\theta_0}^{>\ell}(x_{i,t}).
    \label{eq:base_policy_split}
\end{equation}
We replace $x_{i,t}$ with edited activation $x'_{i,t}$ before the frozen downstream policy is evaluated: $u_{i,t} = G_{\theta_0}^{>\ell}(x'_{i,t})$.
The steering policy does not output a motor command or an action residual
$\Delta u_{i,t}$; it acts only by modifying selected SAE latents
inside the frozen base policy.

% Equivalently,
% \[
%     \pi_{\theta_0}
%     =
%     G_{\theta_0}^{>\ell}
%     \circ
%     F_{\theta_0}^{\leq \ell}.
% \]
% We replace $x_{i,t}$ with edited activation $x'_{i,t}$ before the frozen downstream policy is evaluated:
% \begin{equation}
%     u_{i,t}
%     =
%     G_{\theta_0}^{>\ell}(x'_{i,t}).
%     \label{eq:edited_activation_insert}
% \end{equation}
\subsection{Sparse Activation Basis and Behavior Latent Selection}
\label{sec:sparse_basis}

Editing the dense activation $x_{i,t} \in \mathbb{R}^d$ directly would require the steering policy to jointly control all $d$ of its dimensions (which could be large). Instead, we move to a sparse SAE basis where a few behavior-relevant latents can be edited in isolation while the rest pass through unchanged.
We train an SAE on activations $x_{i,t}$ collected from rollouts of the frozen base policy. The encoder $E_\psi$ maps each activation to a sparse latent vector, and the linear decoder $D_\psi$ reconstructs it back:
\begin{equation}
    \ell_{i,t}=E_\psi(x_{i,t}),
    \qquad
    \hat{x}_{i,t}=D_\psi(\ell_{i,t})=W_D\ell_{i,t}+b_D,
    \qquad
    \ell_{i,t}\in\mathbb{R}^{K_{\mathrm{SAE}}}.
    \label{eq:sae_encode_decode}
\end{equation}
The SAE is trained post-hoc using the standard reconstruction-plus-sparsity objective used for sparse feature decomposition of neural activations
\cite{cunningham2023sparse,bricken2023towards}:
\begin{equation}
    \mathcal{L}_{\mathrm{SAE}}
    =
    \mathbb{E}_{x}
    \left[
    \|x-D_\psi(E_\psi(x))\|_2^2
    +
    \lambda_{\mathrm{sparse}}\|E_\psi(x)\|_1
    \right].
    \label{eq:sae_loss}
\end{equation}

After training the SAE, we use post-hoc probing to select a compact set of behavior-relevant latents for the steering policy to edit. This builds on standard representation probing~\cite{alain2016understanding,kim2018tcav,belinkov2022probing} and SAE feature decomposition~\cite{cunningham2023sparse,templeton2024scaling}. For each target behavior $b$, rollout-derived metrics serve as probes. We score latents using a split-stable absolute Pearson score $S_j^{(b)}=\operatorname{mean}_s|\rho_{j,s}^{(b)}| -\operatorname{std}_s|\rho_{j,s}^{(b)}|$, where $\rho_{j,s}^{(b)}$ is the correlation between latent $j$ and the metric on trajectory split $s$. For multiple metrics, $S_j^{(b)}$ combines these scores. We select the editable latent set as $\mathcal{S}_b=\operatorname{TopK}_{j}S_j^{(b)}$. 
% The SAE basis is behavior-agnostic, while $\mathcal{S}_b$ is behavior-specific (probe details in Appendix~\ref{app:latent_selection}).
% \added{The SAE basis is behavior-agnostic, while $\mathcal{S}_b$ is behavior-specific: velocity probes use the velocity components, formation probes use relational square-geometry quality, and stealth probes use trajectory-modulation metrics such as lateral motion, heading change, and clearance, since camera avoidance is exogenous to the frozen policy's objective and is not directly represented in its activations.}
\added{The SAE basis is behavior-agnostic, while $\mathcal{S}_b$ is behavior-specific: velocity probes use the velocity components, formation probes use relational square-geometry quality, and stealth probes use trajectory-modulation metrics such as lateral motion, velocity, and clearance, since camera avoidance is not directly represented in the frozen policy's activations.} Additional rollout probe metrics are described in
Appendix~\ref{app:latent_selection}.

\subsection{Closed-Loop Affine Activation Editing}
\label{sec:affine_activation_editing}

% Given the selected latent set $\mathcal{S}_b$, we learn a
% behavior-specific steering policy $\pi_{\phi_b}^{\mathrm{steer}}$. At each time step, the steering policy receives a task-specific observation $o^{\mathrm{steer}}_{i,t}$ and outputs affine edit commands for the selected SAE latents:
% \begin{equation}
%     [m_{i,t},c_{i,t}]
%     =
%     \pi_{\phi_b}^{\mathrm{steer}}
%     (o^{\mathrm{steer}}_{i,t}),
%     \qquad
%     m_{i,t},c_{i,t}
%     \in
%     \mathbb{R}^{|\mathcal{S}_b|}.
%     \label{eq:steering_action}
% \end{equation}

Given the selected latent set $\mathcal{S}_b$, we learn a behavior-specific steering policy $\pi_{\phi_b}^{\mathrm{steer}}$. At each time step, the steering policy provides an affine edit, $\xi_{i,t} = (m_{i,t},c_{i,t}) \in \mathbb{R}^{2|\mathcal{S}_b|}$, where $m_{i,t},c_{i,t}\in\mathbb{R}^{|\mathcal{S}_b|}$ are its multiplicative and additive components respectively. 
% applied to the selected SAE latents. 
% The edit is applied before decoding, directly in the SAE latent basis:
% Given the selected latent set $\mathcal{S}_b$, we learn a
% behavior-specific steering policy $\pi_{\phi_b}^{\mathrm{steer}}$. At each time step, the steering policy provides an affine edit
% \begin{equation}
%     \xi_{i,t}
%     =
%     (m_{i,t},c_{i,t})
%     \in
%     \mathbb{R}^{2|\mathcal{S}_b|},
%     \label{eq:latent_edit_command}
% \end{equation}
% where $m_{i,t},c_{i,t}\in\mathbb{R}^{|\mathcal{S}_b|}$ are the multiplicative and additive components of the affine edit applied to the selected SAE latents. 
This edit is applied directly within the SAE latent basis before decoding:
\begin{equation}
    \ell'_{i,t,j}
    =
    \begin{cases}
    m_{i,t,j}\ell_{i,t,j}+c_{i,t,j}, & j\in\mathcal{S}_b,\\
    \ell_{i,t,j}, & j\notin\mathcal{S}_b.
    \end{cases}
    \label{eq:affine_edit}
\end{equation}

Thus, only the behavior-selected SAE latents are modified; all unselected SAE latents are copied unchanged.
The full edited SAE latent is decoded to obtain the edited activation,
$x'_{i,t}=D_\psi(\ell'_{i,t})$, which is then inserted into the frozen policy activation.
When $m_{i,t}=\mathbf{1}$ and $c_{i,t}=\mathbf{0}$, the edited latent reduces to its SAE reconstruction $D_\psi(E_\psi(x_{i,t}))$.

\subsection{Learning Closed-Loop Affine Activation Edits}
\label{sec:learning_steering_policy}

% The affine edit $\xi_{i,t}$ in Eq.~\eqref{eq:affine_edit} is not available as supervised labels. Its effect is only revealed after the edited activation is decoded, inserted into the frozen policy, mapped to a robot action, executed in the simulator, and propagated through future multi-agent states. The useful edit is therefore state-dependent: an edit that helps avoid a camera in one state may move the robot toward an obstacle in another, and an edit that improves formation geometry when agents are dispersed may be harmful when the group is already compact. 
The affine edit $\xi_{i,t}$ is not observed as a supervised target: its effect is only revealed after the edited activation is decoded, passed through the frozen policy, executed as a robot action, and propagated through the multi-agent simulator. Because useful edits depend on the current robot, environment, and team state, we learn them with closed-loop reinforcement learning.
% We therefore cast state-dependent activation editing as a closed-loop reinforcement learning problem.
The affine edit $\xi_{i,t}=(m_{i,t},c_{i,t})$ is the action of the steering policy. 
% It is not a motor command on the robot: it specifies the affine edit applied to the selected SAE latents. The edited activation is then decoded, inserted into the frozen base policy, and the frozen downstream policy produces the robot action $u_{i,t}$.
% This induces a closed-loop transition over the simulator state through the fixed
% base policy and fixed SAE:
% \begin{equation}
%     s_{t+1}
%     \sim
%     P_{\theta_0,\psi}
%     \left(
%     \cdot
%     \mid
%     s_t,
%     \xi_{1:N,t}
%     \right),
%     \label{eq:steering_transition}
% \end{equation}
This induces a closed-loop transition over the simulator state through the fixed base policy and fixed SAE,
$ s_{t+1} \sim P_{\theta_0,\psi}(\cdot \mid s_t,\xi_{1:N,t}) $.
Here, $s_t$ is the simulator state and $\xi_{1:N,t}$ are the affine edits for all robots. The transition kernel $P_{\theta_0,\psi}$ denotes the closed-loop rollout dynamics induced by the simulator, the frozen base policy $\pi_{\theta_0}$, the frozen SAE $(E_\psi,D_\psi)$, and the activation-editing interface from Sec.~\ref{sec:affine_activation_editing}. We do not learn this transition model; the steering policy interacts with it through rollouts. Only the steering-policy parameters $\phi_b$ are optimized. The steering policy maximizes the expected discounted return
\begin{equation}
    J_b(\phi_b)
    =
    \mathbb{E}_{\tau\sim
    (\pi_{\phi_b}^{\mathrm{steer}},P_{\theta_0,\psi})}
    \left[
    \sum_{t=0}^{T}
    \gamma^t
    \frac{1}{N}
    \sum_{i=1}^{N}
    r_{i,t}^{(b)}
    \right],
    \qquad
    \theta_0,\psi \ \mathrm{fixed}.
    \label{eq:steering_return}
\end{equation}
The expectation is over closed-loop rollouts of the steering policy. The base policy parameters $\theta_0$ and SAE parameters $\psi$ remain fixed during both steering-policy training and evaluation. 
% The steering observation contains a task-specific head and a common latent-control tail. 
\added{The steering observation contains a task-specific head and a common latent-control tail containing the selected latents and the previous activation edit.}
% (Appendix~\ref{app:steering_details}). 
All target behaviors use the same activation-editing mechanism but different steering observations and rewards. We use the common reward form
\vspace{-8pt}
% \begin{equation}
%     r_{i,t}^{(b)}
%     =
%     r_{\mathrm{task},i,t}^{(b)}
%     +
%     w_{\mathrm{stab}}^{(b)}
%     r^{\mathrm{stab}}_{i,t}
%     -
%     \lambda_{\mathrm{crash}}^{(b)}
%     C_{i,t}
%     +
%     w_{\mathrm{base}}^{(b)}
%     \tanh
%     \left(
%     \frac{
%     R^{\mathrm{base}}_{i,t}
%     }
%     {
%     \tau_{\mathrm{base}}
%     }
%     \right).
%     \label{eq:reward_template}
% \end{equation}

\begin{equation}
    r_{i,t}^{(b)}
    =
    r_{\mathrm{task},i,t}^{(b)}
    +
    w_{\mathrm{stab}}^{(b)}
    r^{\mathrm{stab}}_{i,t}
    -
    \lambda_{\mathrm{crash}}^{(b)}
    C_{i,t}
    \label{eq:reward_template}
\end{equation}
Here, $r_{\mathrm{task},i,t}^{(b)}$ is the behavior-specific steering objective, $C_{i,t}$ is a weighted crash/contact score, and $r^{\mathrm{stab}}_{i,t}=-\mathds{1}[C_{i,t}>0]$ penalizes unstable or failed states. We optimize Eq.~\eqref{eq:steering_return} with PPO, using a Beta-parameterized policy \cite{petrazzini2021proximal} that produces bounded affine edits (Appendix~\ref{app:steering_details}).

 % We optimize Eq.~\eqref{eq:steering_return} with PPO using a bounded continuous action distribution over $\xi_{i,t}$. In our implementation, this is a Beta-parameterized policy over the affine edit.
 
% details of the action parameterization and bounds are provided in Appendix~\ref{app:steering_details}. 
% PPO is used as the optimizer for the steering policy, while the methodological contribution is the closed-loop latent-edit action space and its insertion into a
% frozen multirobot policy.

%===============================================================================
\section{Experimental Results}
\label{sec:result}
% \vspace{-0.5em}

\begin{figure*}[t]
    \centering
    \includegraphics[width=1.0\linewidth]
    % {corl_2026_template_submission/figures/Result_Main_1.png}
    {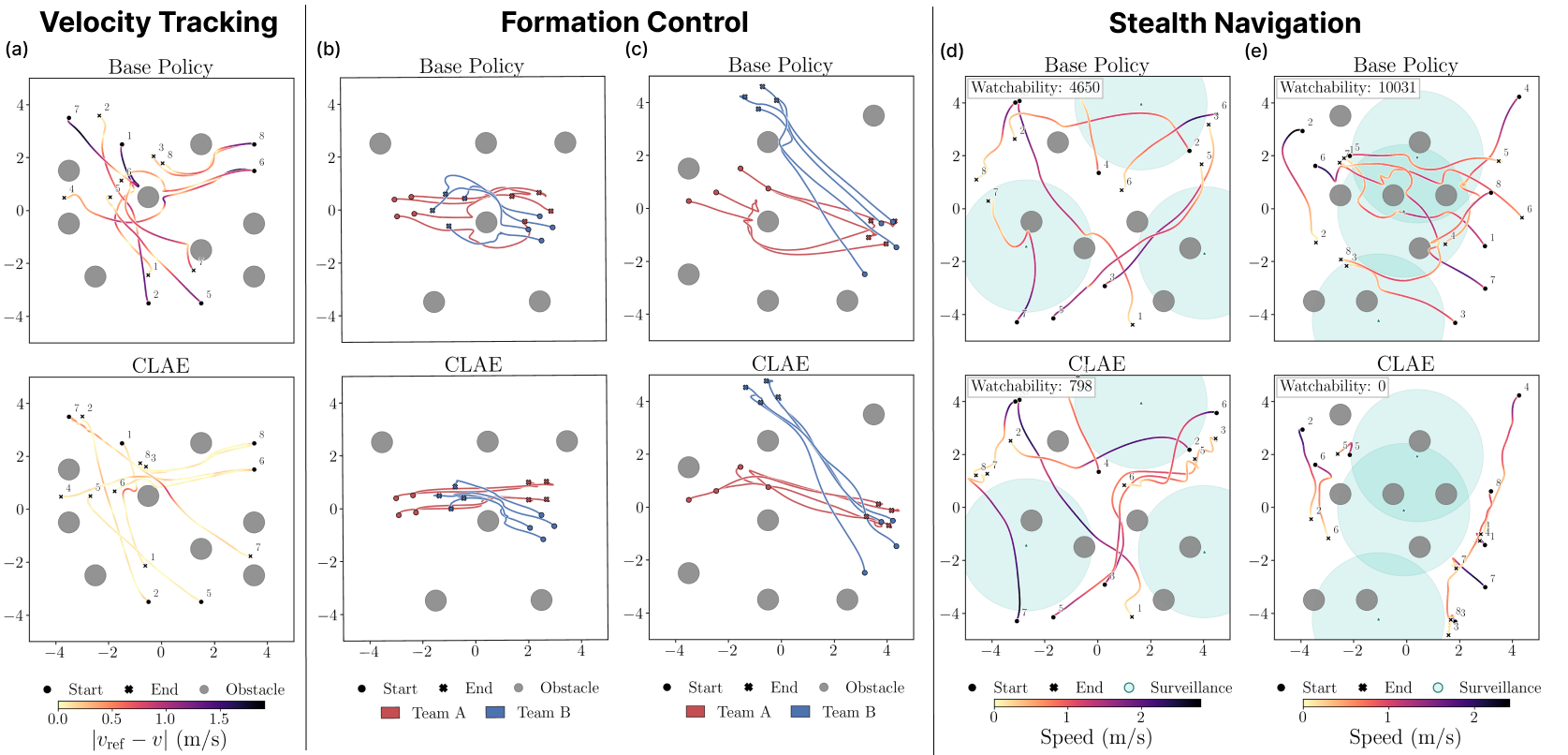}
\caption{\textbf{Qualitative behavior: \method versus the frozen base policy.}
 Trajectory color denotes velocity-tracking error $|v-v^\star|$ in (a), team membership in (b,c), and speed in (d,e). In (b), both the start and goal configurations are square formations; in (c), robots start from an arbitrary configuration and form the target square. (d,e) Stealth navigation. Teal areas denote surveillance areas that must be avoided.
 % \method improves velocity tracking, preserves square formation geometry, and reduces Watchability from $4650\!\to\!798$ in (d) and $10031\!\to\!0$ in (e).
}
% \vspace{-1.5em}
\label{fig:clae_qualitative}
\end{figure*}

\subsection{Experimental Setup}
\label{sec:setup}
% \vspace{-0.5em}

% \textbf{Simulator \& Base Policy.}
% We use the QuadSwarm~\cite{huang2023quadswarm} simulation environment with 8 quadrotors navigating a $10\times 10\times 10$\,m room containing static cylindrical obstacles. % \textbf{Base policy.} 
% We adopt the end-to-end, decentralized RL policy of Huang et al.~\cite{huang2024collision} as the base controller. Each robot observes its own state and goal, the relative states of its two nearest neighbors, and a $3\times 3$ signed-distance field encoding of nearby obstacles. These observations are processed by three separate MLP encoders. The neighbor and obstacle embeddings are fused through a multi-head attention module, concatenated
% with the self embedding, and passed through downstream MLP layers and an action head that outputs four normalized rotor thrusts. 

\textbf{Simulator \& Base Policy:}
We use the QuadSwarm~\cite{huang2023quadswarm} simulator with 8 quadrotors navigating a $10\times 10\times 10$\,m room with static cylindrical obstacles. We adopt the end-to-end, decentralized RL policy~\cite{huang2024collision} as the base controller. Each robot observes its state and goal, the relative states of its two nearest neighbors, and a $3\times 3$ signed-distance field of nearby obstacles. Three separate MLPs encode these observations. Neighbor and obstacle embeddings are fused via multi-head attention, concatenated with the self embedding, and passed through downstream MLPs to an action head outputting four normalized rotor thrusts.

\textbf{Intervention point:} CLAE requires only white-box intermediate activation access; the SAE trains on the intercepted block, and probing selects behavior-relevant latents. We intercept the first fused activation $x_{i,t}$ post-attention, as it is the earliest representation combining self, neighbor, and obstacle information. \added{All activation-editing baselines and ablations use this same activation,
so observed differences reflect the editing mechanism rather than layer choice.}

\subsection{Behaviors}
\label{sec:tasks}

We steer the pretrained multi-robot navigation policy towards the following behaviors: \textbf{Velocity-profile tracking.} Each robot is assigned a reference velocity profile $v^\star_{i,t}$ and must track it while still reaching its goal and avoiding obstacles. The reference is generated by a minimum snap planner~\cite{mellinger2011minimum}, independent of the learned policy and made feasible in the obstacle-rich environment using a CBF-based~\cite{wang2016safety} filter.
% Because the reference varies over time and across robots, this task tests whether activation editing can induce precise time-varying motion, beyond the coarse faster-or-slower behavior changes demonstrated in prior single-robot activation steering. 
% \textbf{Velocity-profile tracking.}
% Each robot receives a reference velocity profile $v^\star_{i,t}$ and must track it while reaching its goal and avoiding obstacles. The reference is generated by a nominal planner~\cite{mellinger2011minimum}, independent of the learned policy, and made feasible in clutter using CBF-based post-processing~\cite{wang2016safety}. Since the reference varies over time and across robots, this task tests whether activation editing can induce precise time-varying motion beyond the coarse faster-or-slower changes shown in prior single-robot activation steering. 
\textbf{Multi-robot formation control.}
The robots are partitioned into two four-robot groups, and each group is assigned a target formation. We use a square with desired side length $s^\star=0.5$\,m; other formations can be specified by changing the reference distance pattern. 
% This task tests whether activation edits can coordinate multi-robot behavior, since the desired edit for each robot depends on teammate states rather than only its own state.
\textbf{Camera-aware stealth navigation.}
We introduce surveillance cameras that are \emph{absent} from the base policy's original training. Each environment contains three cameras with surveillance radius $2.5$\,m, placed to cover large portions of the free space. The steering policy receives camera-risk features and must reduce exposure while preserving goal reaching and collision avoidance.

All behaviors use the reward form in Eq.~\eqref{eq:reward_template}, with only the task reward changing across behaviors:
\begin{equation}
\begin{split}
    r^{\text{vel}}_{\text{task},i,t} &= w_{\text{track}} r^{\text{track}}_{i,t},\quad \quad \quad
    r^{\text{form}}_{\text{task},i,t} = w_{\text{shape}} r^{\text{shape}}_{i,t}
      + w_{\text{goal}} r^{\text{goal}}_{i,t} \\
    r^{\text{stealth}}_{\text{task},i,t} &= w_{\text{zone}} r^{\text{zone}}_{i,t}
      + w_{\text{margin}} r^{\text{margin}}_{i,t}
      + w_{\text{goal}}^{\text{stealth}} r^{\text{goal}}_{i,t}
\end{split}
\label{eq:task_reward_summary}
\end{equation}
% $r^{\text{vel}}_{\text{task},i,t}$ rewards reference tracking, $r^{\text{form}}_{\text{task},i,t}$ rewards formation geometry and goal progress, and $r^{\text{stealth}}_{\text{task},i,t}$ primarily rewards reduced camera exposure, with a small goal-progress term to bias the edited policy toward goal reaching. 
% Full observation heads, action bounds, reward definitions, and hyperparameters are provided in Appendix~\ref{app:steering_details}. 
\added{Here $r^{\text{track}}_{i,t}$ is the negative tracking error against the reference velocity, $r^{\text{shape}}_{i,t}$ penalizes deviation from the target square geometry, and $r^{\text{goal}}_{i,t}$ penalizes normalized distance to the assigned goal. The stealth terms combine a binary penalty $r^{\text{zone}}_{i,t}$ for entering a camera's valid surveillance footprint with a soft margin $r^{\text{margin}}_{i,t}$ that grows as a robot approaches a camera center, giving PPO a dense signal around an otherwise sparse objective; exposure remains the primary stealth objective, with $w^{\text{stealth}}_{\text{goal}}$ small and serving only to bias the edited policy toward goal reaching.} The task-reward definitions are detailed in
Appendix~\ref{app:task_rewards}.
\added{Each steering policy conditions on the selected SAE latents and its previous activation edit, together with a compact task-specific head: velocity tracking receives reference-tracking information; formation receives teammate-relative information; and stealth receives local surveillance-camera information. All steering inputs are available at inference time from the robot state estimate, the task specification, and the frozen-policy activation stream.
}
\begin{table}[t]
\centering
\small
\setlength{\tabcolsep}{4pt}
\renewcommand{\arraystretch}{1.12}
\resizebox{\linewidth}{!}{%
\begin{tabular}{l ccc l}
\toprule
\textbf{Method}
& \textbf{SAE}
& \textbf{RL-optimized}
& \textbf{State-}
& \textbf{Intervention rule} \\
&
& \textbf{intervention}
& \textbf{conditioned edit}
& \\
\midrule

Base Policy
& --- & --- & ---
& $x'_t = x_t$ \\

\addlinespace[2pt]

DSRL (distilled diffusion policy) \cite{wagenmaker2025dsrl}
& --- & \checkmark & \checkmark
& $a_k = D(s_k,w_k),\; w_k \sim \pi^{\mathrm{noise}}(\cdot \mid s_k)$ \\

Residual Action RL \cite{johannink2018residual}
& --- & \checkmark & \checkmark
& $a_t = u_t + \Delta u_t(s_t)$ \\

\addlinespace[2pt]

Fixed Scalar SAE Steering \cite{swann2026sparse}
& \checkmark & --- & ---
& $x'_t = x_t + \alpha d_j$ \\

Linear Probe Intervention \cite{buurmeijer2026observing}
& --- & --- & \checkmark
& $x'_t = x_t + \mathrm{clip}_\eta(u_t)$ \\

Closed-Loop Multiplicative Editing
& \checkmark & \checkmark & \checkmark
& $\ell'_{S,t} = m_t \odot \ell_{S,t}$ \\
Closed-Loop Additive-Only Editing
& \checkmark & \checkmark & \checkmark
& $\ell'_{S,t} = \ell_{S,t} + c_t$ \\
\textbf{\method (ours)}
& \checkmark & \checkmark & \checkmark
& $\ell'_{S,t} = m_t \odot \ell_{S,t} + c_t$ \\

\bottomrule
\end{tabular}%
}
\caption{\addedv{\textbf{Compared methods.}
All steering methods keep their respective base policies fixed and differ
in where and how the intervention is produced.}}
\label{tab:methods}

\end{table}

%%%%%%%%%%%%%%%%%

\subsection{Baselines and Evaluation Protocol}
\label{sec:methods}
% \vspace{-0.5em}

% \textbf{Baselines.} We compare \method\ against the frozen base policy, two activation editing baselines, and a multiplicative-only ablation, all using the same intervention layer and frozen downstream policy (Appendix~\ref{sec:baseline_details}, Table~\ref{tab:methods}). Since existing robotic learning activation-steering methods are not designed for closed-loop multi-robot behavior steering, we adapt the closest single-robot VLA-style interventions: fixed SAE decoder-direction steering~\cite{swann2026sparse} and linear-probe intervention~\cite{buurmeijer2026observing}. We evaluate these on velocity tracking and formation control, but not for stealth navigation because the stealth objective is exogenous to the frozen policy's training objective and does not correspond to a behavior variable that is directly represented or can be readily linearly probed in the base policy activation space. We also evaluate \textbf{Closed-Loop Multiplicative Editing}, which removes the additive offset from Eq.~\eqref{eq:affine_edit} and applies only $\ell'_{S,t}=m_t \odot \ell_{S,t}$, isolating the contribution of the full affine edit. 

\textbf{Baselines.} \addedar{We compare \method\ against the frozen base policy,
two non-activation interventions, two prior activation-editing
baselines, and two single-component ablations of \method\ (Table~\ref{tab:methods}).}
\added{All steering methods keep their respective base policies fixed; the activation-editing methods additionally share the same intervention layer
and frozen downstream policy.}
% \added{\textbf{DSRL}~\cite{wagenmaker2025dsrl} steers the initial noise of  frozen denoising policy; our base policy is feedforward and admits no such intervention. Applying it therefore requires distilling the base policy into a diffusion policy, which we train by behavior cloning on base-policy rollouts. We then train a per-behavior latent-noise actor on the frozen clone, and report the unsteered clone to separate distillation loss from steering effects.} 
\addedv{\textbf{DSRL}~\cite{wagenmaker2025dsrl} is included as an auxiliary diffusion-policy comparison. Since our base controller has no diffusion-noise interface, we train a diffusion policy~\cite{chi2023diffusion} by behavior cloning on
% $10^4$ 
base policy rollouts, freeze it, and adapt DSRL to steer its initial noise for each target behavior. Because behavior cloning can incur compounding errors under policy-induced distribution shift~\cite{ross2011reduction,loquercio2021agile}, a relevant consideration in our closed-loop multi-robot setting, we report both the unsteered and steered diffusion policies to distinguish the pre-steering performance gap from steering gains.}
\added{\textbf{Residual Action RL}~\cite{johannink2018residual,silver2018residual} learns a bounded corrective term on the frozen base policy's output using the same steering observations, reward, RL algorithm, and environment-step budget as \method, isolating intervention in action space from intervention in activation space.} Since existing robotic learning activation-steering methods are not designed for closed-loop multi-robot behavior steering, we adapt the closest single-robot VLA-style interventions: \added{\textbf{Fixed Scalar SAE Steering}~\cite{swann2026sparse}, which shifts the activation along one SAE decoder direction, and \textbf{Linear Probe Intervention}~\cite{buurmeijer2026observing}, which fits a probe offline and applies a closed-form perturbation, so its edit varies with state but is not optimized from closed-loop return.} 
% We evaluate these on velocity tracking and formation control, but not on stealth navigation, because the stealth objective is exogenous to the frozen policy's training objective and does not correspond to a behavior variable that is directly represented or readily linearly probed in the base policy activation space. 
\added{We evaluate these on velocity tracking and formation control but not stealth navigation, since the stealth objective is exogenous to the frozen policy's training objective and is not directly represented in its activations.}
% We also evaluate \textbf{Closed-Loop Multiplicative Editing}, which removes the additive offset from Eq.~\eqref{eq:affine_edit} and applies only $\ell'_{S,t}=m_t \odot \ell_{S,t}$, isolating the contribution of the affine edit.
\addedar{We also evaluate two single-component ablations of
Eq.~\eqref{eq:affine_edit} across all three behaviors.
\textbf{Closed-Loop Multiplicative Editing} removes the additive term by
setting $c_t=\mathbf{0}$, yielding
$\ell'_{S,t}=m_t\odot\ell_{S,t}$.
\textbf{Closed-Loop Additive-Only Editing} instead fixes
$m_t=\mathbf{1}$, yielding
$\ell'_{S,t}=\ell_{S,t}+c_t$.
Both retain learned, state-dependent closed-loop steering while restricting
the edit to a single affine component. Comparing these variants with
\method tests the contribution of jointly learning multiplicative and
additive edits. }
% Implementation details are provided in Appendix~\ref{sec:baseline_details}.}

% the full affine edit.

% All methods act at the same intermediate activation, and none modify the base policy weights or add an action-residual controller; they differ only in the edit form, summarized in Table~\ref{tab:methods}. 
% Following the evaluation protocol of LAE~\cite{das2025latent}, we run all quantitative
% comparisons in a deterministic simulator setting: identical initial
% conditions, goals, and obstacle configurations produce identical unedited
% trajectories. This removes evaluation noise from motor noise, sensor noise,
% and policy action sampling, allowing observed differences between methods
% to be attributed to the activation editing itself rather than to stochastic
% variation.
% Following LAE~\cite{das2025latent}, we evaluate all methods in a deterministic simulator setting with matched initial conditions, goals, and obstacle configurations, so differences can be attributed to the activation-editing mechanism rather than stochastic rollout variation.

\textbf{Metrics.}
% All metrics are averaged over $1000$ episodes per method.
\emph{Behavior-specific:}
\textbf{Velocity Error (Vel.\ Err.)} (m/s) is the mean absolute
component-wise error between each robot's velocity and its reference,
$|v - v^\star|$, averaged over axes, robots, and timesteps;
$|e_x|, |e_y|, |e_z|$ are its per-axis components.
% \textbf{Formation Error (Form.\ Err.)} measures how closely each four-robot
% group maintains the target square formation, with $0$ corresponding to
% perfect target geometry (full definition in Appendix~B).
\added{\textbf{Formation Error (Form.\ Err.)} measures how closely each four-robot
group maintains the target square formation, as the mean relative deviation of
the group's six pairwise XY distances from their target values in the square
(four sides and two diagonals); $0$ corresponds to perfect target geometry.}
\textbf{Watchability} is the total number of robot-camera visibility events
per episode, summed over robots, cameras, and timesteps; $0$ means no robot
was ever visible to any camera.
\emph{General:}
\textbf{Collision-Free Rate (CF)} (\%) is the fraction of episodes with no
robot-robot or robot-obstacle collisions;
% \textbf{Success Rate (SR)} is the fraction of robots that reach the goal
% safely, defined as ending within $0.8$\,m of the assigned goal in a
% collision-free episode;
\added{\textbf{Success Rate (SR)} is the fraction of robots that reach the goal
safely, defined as ending within $0.8$\,m of the assigned goal without
being involved in a collision;}
\textbf{Goal Distance (Goal Dist.)} (m) is the mean final distance
between each robot and its assigned goal.

\begin{table*}[t]
\centering
\small
\setlength{\tabcolsep}{4pt}
\renewcommand{\arraystretch}{0.98}
\begin{tabular}{l cccc ccc}
\toprule
\textbf{Method}
& \multicolumn{4}{c}{\textit{Behavior-specific}\,$\downarrow$}
& \multicolumn{3}{c}{\textit{General}} \\
\midrule

\textit{(a) Velocity-Profile Tracking}
  & \textbf{Vel.\ Err.} & $|e_x|$ & $|e_y|$ & $|e_z|$
  & CF\,(\%)\,$\uparrow$ & SR\,$\uparrow$ & Goal\,(m)\,$\downarrow$ \\
\cmidrule(lr){2-5}\cmidrule(lr){6-8}

Base Policy
  & 0.322 & 0.221 & 0.212 & 0.149
  & 99.0 & 0.888 & 0.457 \\
\addlinespace[3pt]

Distilled Diffusion Policy (unsteered)
  & 0.340 & 0.234 & 0.223 & 0.156
  & 48.0 & 0.817 & 0.496 \\

DSRL (distilled diffusion policy) \cite{wagenmaker2025dsrl}
  & 0.298 & 0.218 & 0.216 & 0.147
  & 18.7 & 0.476 & 0.847 \\

Residual Action RL \cite{johannink2018residual}
  & 0.269 & 0.172 & 0.174 & 0.080
  & 87.7 & 0.001 & 3.170 \\
\addlinespace[3pt]

Fixed Scalar SAE Steering \cite{swann2026sparse}
  & 0.305 & 0.200 & 0.189 & 0.134
  & 99.0 & 0.882 & 0.559 \\

Linear Probe Intervention \cite{buurmeijer2026observing}
  & 0.294 & 0.189 & 0.185 & 0.139
  & 99.0 & 0.836 & 0.593 \\
\addlinespace[3pt]

Closed-Loop Multiplicative Editing
  & 0.318 & 0.214 & 0.219 & 0.134
  & 98.8 & 0.861 & 0.501 \\

\addedar{Closed-Loop Additive-Only Editing}
  & 0.129 & 0.104 & 0.084 & 0.059
  & 80.6 & 0.857 & 1.195 \\
\addlinespace[3pt]

\textbf{\method (ours)}
  & \textbf{0.085} & \textbf{0.075} & \textbf{0.057} & \textbf{0.039}
  & 98.7 & 0.899 & 0.456 \\

\midrule

\textit{(b) Multi-Robot Formation Control}
  & \multicolumn{4}{c}{\textbf{Form.\ Err.}}
  & CF\,(\%)\,$\uparrow$ & SR\,$\uparrow$ & Goal\,(m)\,$\downarrow$ \\
\cmidrule(lr){2-5}\cmidrule(lr){6-8}

Base Policy
  & \multicolumn{4}{c}{0.280}
  & 99.5 & 0.997 & 0.271 \\
\addlinespace[3pt]

Distilled Diffusion Policy (unsteered)
  & \multicolumn{4}{c}{0.297}
  & 79.6 & 0.793 & 0.293 \\

DSRL (distilled diffusion policy) \cite{wagenmaker2025dsrl}
  & \multicolumn{4}{c}{0.241}
  & 70.2 & 0.682 & 0.428 \\

Residual Action RL \cite{johannink2018residual}
  & \multicolumn{4}{c}{0.247}
  & 94.6 & 0.944 & 0.498 \\
\addlinespace[3pt]

Fixed Scalar SAE Steering \cite{swann2026sparse}
  & \multicolumn{4}{c}{0.466}
  & 22.3 & 0.000 & 3.713 \\

Linear Probe Intervention \cite{buurmeijer2026observing}
  & \multicolumn{4}{c}{0.315}
  & 95.8 & 0.972 & 0.483 \\
\addlinespace[3pt]

Closed-Loop Multiplicative Editing
  & \multicolumn{4}{c}{0.276}
  & 99.2 & 0.989 & 0.242 \\

\addedar{Closed-Loop Additive-Only Editing}
  & \multicolumn{4}{c}{0.096}
  & 76.4 & 0.814 & 0.291 \\
\addlinespace[3pt]

\textbf{\method (ours)}
  & \multicolumn{4}{c}{\textbf{0.086}}
  & 98.5 & 0.994 & 0.281 \\

\midrule

\textit{(c) Camera-Aware Stealth Navigation}
  & \multicolumn{4}{c}{\textbf{Watchability}}
  & CF\,(\%)\,$\uparrow$ & SR\,$\uparrow$ & Goal\,(m)\,$\downarrow$ \\
\cmidrule(lr){2-5}\cmidrule(lr){6-8}

Base Policy
  & \multicolumn{4}{c}{3596.6}
  & 99.1 & 0.924 & 0.530 \\
\addlinespace[3pt]

Distilled Diffusion Policy (unsteered)
  & \multicolumn{4}{c}{3755.6}
  & 27.3 & 0.245 & 0.560 \\

DSRL (distilled diffusion policy) \cite{wagenmaker2025dsrl}
  & \multicolumn{4}{c}{2614.2}
  & 12.4 & 0.101 & 0.733 \\

Residual Action RL \cite{johannink2018residual}
  & \multicolumn{4}{c}{3636.3}
  & 43.7 & 0.284 & 0.939 \\
\addlinespace[3pt]

Closed-Loop Multiplicative Editing
  & \multicolumn{4}{c}{3453.2}
  & 97.0 & 0.897 & 0.604 \\

\addedar{Closed-Loop Additive-Only Editing}
  & \multicolumn{4}{c}{1261.0}
  & 45.8 & 0.350 & 1.290 \\
\addlinespace[3pt]

\textbf{\method (ours)}
  & \multicolumn{4}{c}{\textbf{1151.3}}
  & 96.5 & 0.912 & 0.552 \\

\bottomrule
\end{tabular}
\caption{\textbf{Behavior-steering results across three tasks.}
\added{For each behavior, all methods are averaged over the same 1000 distinct environment configurations. Activation-editing and residual-action methods act on the base RL
policy, whose weights remain fixed during steering. The unsteered diffusion
policy and DSRL rows use a separately distilled diffusion policy.}}
\label{tab:main_results}
\end{table*}

\subsection{Results} 
\label{sec:main_results}
% \satya{Better Section Name}
% \vspace{-0.5em}

Table~\ref{tab:main_results} summarizes the quantitative behavior-steering results and Figure~\ref{fig:clae_qualitative} shows the corresponding qualitative behavior changes in representative rollouts. 
% \added{All reported results are averaged over 1000 differently seeded environment configurations.}
\added{For each behavior, results for all methods are averaged over the same set of 1,000 distinct environment configurations, each generated using a different random seed.}
% Together, they show that \method improves the target behavior while preserving the frozen base policy's navigation competence.
% Table~\ref{tab:main_results} reports quantitative results across all three tasks, and Figure~\ref{fig:clae_qualitative} visualizes representative rollouts. Across velocity tracking, formation control, and stealth navigation, \method attains the best task-specific metric while preserving the base policy's core navigation behavior.  

% while keeping collision-free and goal-reaching performance close to the frozen base policy.

\begin{figure}[t]
  \centering
    \includegraphics[width=0.95\linewidth]{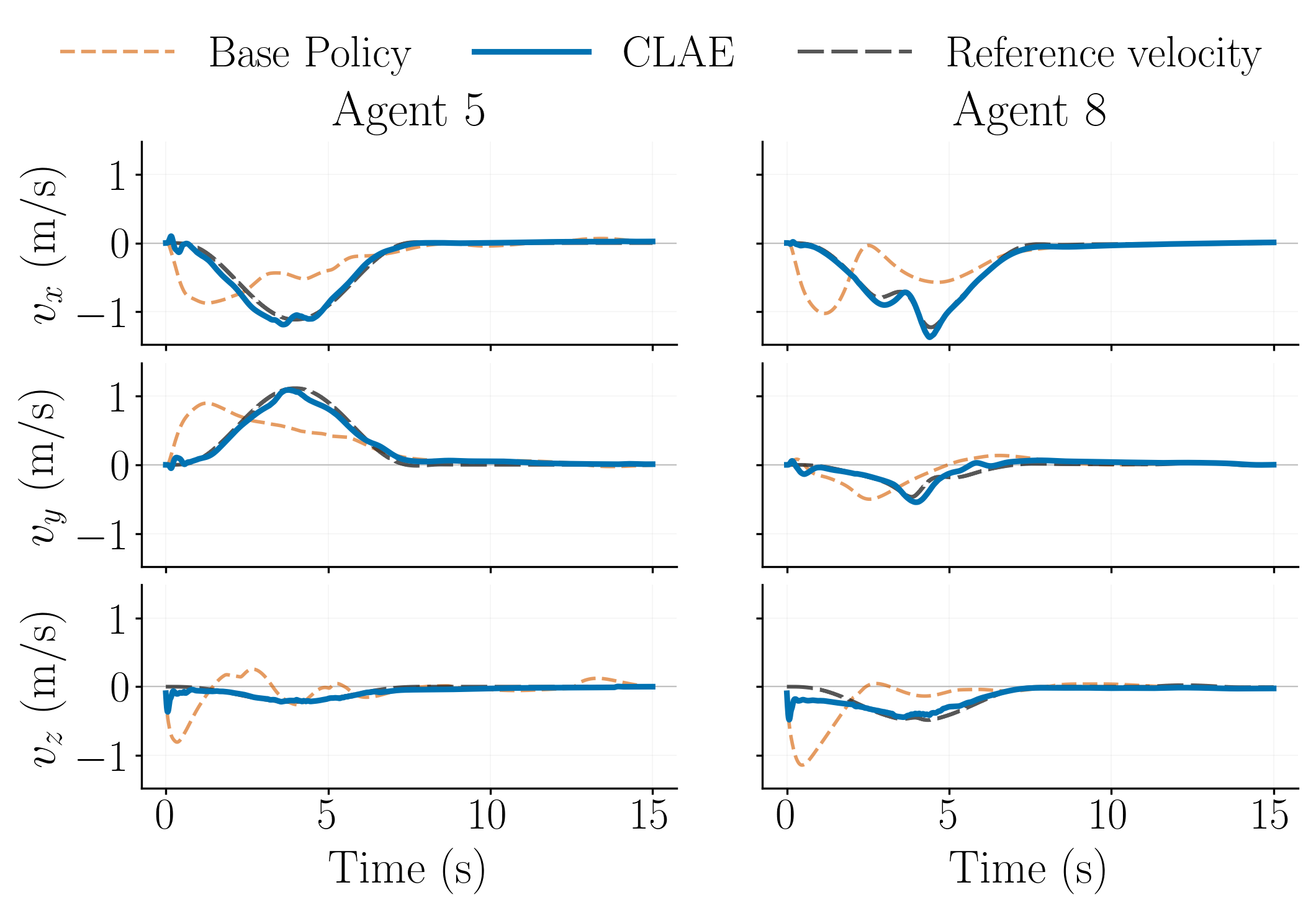}
\caption{\textbf{Per-axis velocity tracking.} \method (blue) follows the reference (dashed grey) on $v_x$, $v_y$, $v_z$ for two representative robots; the unedited base policy (dashed orange) diverges on every axis.}
  \label{fig:velocity_components}
  \vspace{-1.0em}
\end{figure}

\textbf{Velocity-profile tracking.}
\method reduces the mean velocity-tracking error from $0.322$\,m/s (base policy) to $0.085$\,m/s, a $\sim$$3.8\times$ improvement, with consistent
per-axis gains on $|e_x|$, $|e_y|$, and $|e_z|$
(Table~\ref{tab:main_results}a). 
% This gap is expected, as the base policy was
% never trained to track an external velocity signal, and it is precisely the gap that activation editing must close. 
The qualitative result in Figure~\ref{fig:clae_qualitative}a confirms the effect at the trajectory
level: \method trajectories, coloured by tracking residual $|v - v^\star|$, remain uniformly pale, whereas the base-policy trajectories show large residuals throughout. 
Figure~\ref{fig:velocity_components}
shows the per-axis velocity traces for two representative robots: \method follows the reference profile closely on $v_x$, $v_y$, and $v_z$, while the base policy produces its own goal-driven velocity profile that diverges sharply from the reference. 
% Fixed Scalar SAE Steering and Linear Probe
% Intervention reduce tracking error only marginally ($\leq 10\%$): a fixed-direction scalar shift or a closed-form linear perturbation struggles to follow a time-varying reference profile across the full episode.
% The prior-style baselines improve tracking  marginally, and the multiplicative-only variant does not improve over the base policy; state-dependent affine edits are needed to track a time-varying reference rather than change speed along a fixed direction.
% \added{The non-activation interventions reduce tracking error only to $0.238$ and $0.269$, still far above \method's $0.085$. The activation-editing baselines improve tracking marginally, and the multiplicative-only variant does not improve over the base policy; state-dependent affine edits are needed to track a time-varying reference rather than change speed along a fixed direction.}
\added{Residual Action RL, acting on the same frozen base policy, reduces tracking error to $0.269$, still far above \method's $0.085$. The activation-editing baselines improve tracking marginally, while the multiplicative-only variant remains close to the base policy}.
\addedar{Additive-only editing is substantially more effective, reaching
$0.129$\,m/s, but reduces the collision-free rate to $80.6\%$.
Together, these results support the benefit of state-dependent affine
edits for tracking a time-varying reference while retaining the base
policy's navigation performance.}

% The multiplicative ablation provides essentially no
% improvement: gain modulation alone cannot track the reference, and the small additive offset is needed in combination with multiplicative transformation to close the gap. 
% Tracking arbitrary references suggests activation editing can adapt robot motion to any deployment-time preference, extending beyond our evaluated examples.

% The ability to track an \emph{arbitrary
% feasible} reference profile suggests that activation editing could reshape individual robot motion toward any deployment-time
% motion preference, not only the specific references evaluated here.\satya{compress the last statement}

\textbf{Multi-robot formation control.}
\method reduces formation error from $0.280$ (base policy) to $0.086$, a
$\sim$$3.2\times$ improvement, while maintaining $98.5\%$ collision-free
episodes and $99.4\%$ robot-success
(Table~\ref{tab:main_results}b). Figure~\ref{fig:clae_qualitative}b
and~\ref{fig:clae_qualitative}c show two representative configurations. In (b), where the start and goal are both
square-structured, the base policy lets the two four-robot groups drift apart near the obstacle, while \method preserves a tight square formation geometry during
navigation. In (c), the starting configuration is arbitrary and only the goal forms a square; \method pulls the robots into formation \emph{en route} and arrives in the
target geometry, whereas the base policy reaches goals without maintaining team structure. 
% The prior-style baselines struggle on this team-dependent objective, while the multiplicative-only variant barely changes to the base policy formation behavior, indicating that formation steering requires affine edits that vary with teammate configuration.
% \added{The non-activation interventions improve formation error only to $0.218$ and $0.247$, against $0.086$ for \method under the same reward and environment-step budget. The activation-editing baselines also struggle on this team-dependent objective, and the multiplicative-only variant barely changes the base policy's formation behavior, indicating that formation steering requires affine edits that vary with teammate configuration.}
\added{Residual Action RL improves formation error to $0.247$, compared with $0.086$ for \method under the same reward and environment-step budget. The activation-editing baselines also struggle on this team-dependent objective, while the multiplicative-only variant remains close to the base policy.
% supporting the benefit of state-dependent affine edits that vary with teammate configuration.
}
 \addedar{The additive-only ablation reaches a
formation error of $0.096$, close to \method's $0.086$, but its
collision-free and robot-success rates fall to $76.4\%$ and $81.4\%$,
respectively, compared with $98.5\%$ and $99.4\%$ for \method.  This highlights the benefit of the
full affine edit for inducing team-dependent formation behavior while
preserving navigation competence.}
% Fixed Scalar SAE Steering collapses on this task ($22.3\%$
% collision-free, $0\%$ goal success): a single state-independent direction cannot encode a team-dependent objective whose correct edit varies with teammate positions. Linear Probe Intervention degrades both
% safety and goal progress; while its intervention is recomputed at each step, the closed-form linear update is not optimized for the team-dependent target geometry. The multiplicative ablation barely changes the base policy's formation behavior, again indicating that gain
% modulation alone is not enough and a small additive component is needed alongside the multiplicative gain. 

\textbf{Camera-aware stealth navigation.}
\method reduces the Watchability score from $3596.6$ to $1151.3$, a $\sim$$3.1\times$ reduction in cumulative camera exposure, while maintaining a $96.5\%$ collision-free rate and a $91.2\%$ robot-success rate (Table~\ref{tab:main_results}c). 
\addedar{The single-component ablations reveal complementary limitations:
multiplicative-only editing reduces exposure by only about $4\%$, indicating
limited camera-aware steering from gain modulation alone, whereas
additive-only editing substantially reduces Watchability to $1261.0$ but
lowers the collision-free rate to $45.8\%$. Thus, additive offsets alone
can improve the target behavior, but at a substantial cost to the base
policy's navigation performance. The full affine edit, \method, avoids this
tradeoff, achieving the strongest exposure reduction while largely retaining
the underlying navigation behavior.}

In Fig.~\ref{fig:clae_qualitative}d, where low-exposure goal-reaching paths exist, \method routes robots around camera footprints or drives them quickly through exposed regions, reducing \emph{cumulative} rather than instantaneous exposure. In Fig.~\ref{fig:clae_qualitative}e, where camera coverage is denser, \method prioritizes exposure reduction over goal reaching, reducing Watchability to $0$ while some robots stop short of their goals. Together, these examples show that the same activation-editing interface can discover qualitatively different deployment-time strategies, trading off goal progress and stealth behavior.

% These two regimes illustrate that the steering policy is expressive enough to discover qualitatively different deployment-time strategies under the same activation-editing interface, trading off goal progress and exposure depending on the geometry of the
% environment.

% Finally, DSRL improves all three behavior-specific metrics relative to the
% unsteered distilled diffusion policy, but both variants retain collision-free
% rates well below those of the original base policy across all three tasks.
% We attribute this navigation degradation primarily to distillation-induced
% covariate shift~\cite{czarnecki2019distilling}; steering does not recover
% the lost navigation performance here, likely because the support of the
% distilled policy's action distribution is too narrow. Because DSRL operates
% on a separately distilled diffusion policy, we treat it as an auxiliary
% comparison rather than a controlled comparison on a shared base policy.

Finally, DSRL improves all three behavior-specific metrics relative to the
unsteered distilled diffusion policy, but both variants retain collision-free
rates well below those of the original base policy across all three tasks.
We attribute the navigation degradation primarily to distillation-induced
covariate shift~\cite{czarnecki2019distilling}. DSRL is unable to recover
the lost navigation performance from the distilled policy. We hypothesize this is due to the support of the distilled policy's action distribution being too narrow. Because DSRL operates
on a separately distilled diffusion policy, we treat it as an auxiliary
comparison rather than a controlled comparison on a shared base policy.

\textbf{Summary.}
Across all three tasks, \method steers the frozen base policy toward behaviors it was never trained to primarily express, per-robot velocity tracking,
multi-robot formation control, and camera-aware stealth navigation, while keeping collision avoidance and goal reaching behavior close to the unedited base policy. 
% \addedv{Compared with Residual Action RL under the same base policy, task reward, RL algorithm, and environment-step budget, \method achieves lower velocity error, formation error, and camera exposure.
% % Noise steering improves all three behavior-specific metrics relative to the unsteered distilled diffusion policy, but its collision-free rates remain well below those of the
% % base policy. Further, we see that both DSRL and the distilled diffusion policy baselines show poor performance (on all three tasks). We attribute this to distillation: the distilled policy is susceptible to covariate shift~\cite{czarnecki2019distilling}, which steering alone cannot remedy here, as we believe the support of the action distribution is too narrow. 
% DSRL improves all three behavior-specific metrics relative to its
% unsteered distilled diffusion policy, but both exhibit poor general
% navigation performance across all three tasks, with collision-free rates
% well below those of the original base policy. We attribute this
% navigation degradation primarily to distillation: the distilled policy
% is susceptible to covariate shift~\cite{czarnecki2019distilling}, which
% steering alone cannot remedy here, likely because the support of its
% action distribution is too narrow.
% }

These results show that small, bounded affine edits at each policy step can compound over closed-loop execution to produce substantial behavior-level steering.

\begin{figure*}[h]
    \centering
    \includegraphics[width=0.90\linewidth]{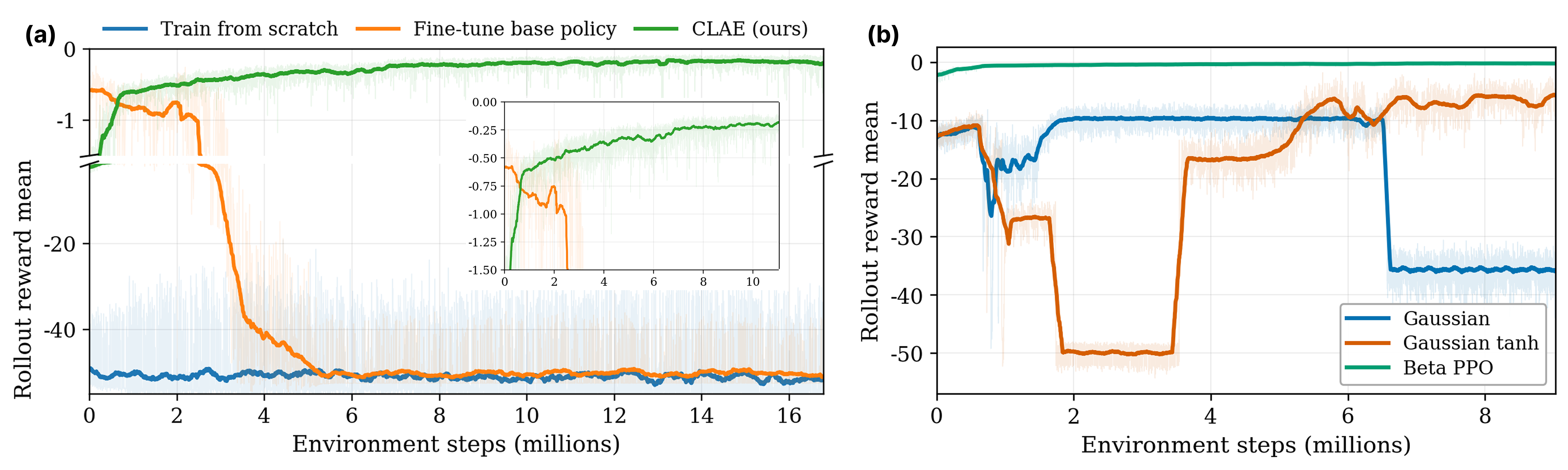}
    \caption{\textbf{Training dynamics for design ablations.}
    \textbf{(a)} \method versus fine-tuning and training from scratch on the
    formation objective. \textbf{(b)} Beta, Gaussian, and Gaussian-tanh PPO
    action heads. Curves are from representative runs and smoothed for
    visualization; final-checkpoint rollout performance is reported in
    Table~\ref{tab:ablation_formation}.}
    \label{fig:ablation}

\end{figure*}

\begin{table}[t]
\centering
\setlength{\tabcolsep}{4pt}
\renewcommand{\arraystretch}{1.1}
\resizebox{\columnwidth}{!}{%
\begin{tabular}{l c c c c}
\toprule
\textbf{Method}
  & \textbf{Form.\ Err.}\,$\downarrow$
  & CF\,(\%)\,$\uparrow$
  & SR\,$\uparrow$
  & Goal\,(m)\,$\downarrow$ \\
\midrule
Base Policy                          & 0.280 & 99.5 & 0.997 & 0.271 \\
\added{SAE reconstruction (no steering)}     & 0.284 & 99.1 & 0.990 & 0.275 \\
\addlinespace[4pt]
Fine-tune base policy                & 0.199 & 93.5 & 0.890 & 0.199 \\
Train from scratch                   & 4.715 & 0.0  & 0.000 & --- \\
\addlinespace[4pt]
\method w/ Gaussian PPO              & 0.761 & 2.6  & 0.000 & 0.761 \\
\method w/ Gaussian-tanh PPO         & 3.276 & 0.0  & 0.000 & --- \\
\addlinespace[4pt]
\method w/o SAE (full-activation)    & 0.797 & 66.6 & 0.127 & 0.797 \\
% \addlinespace[4pt]
\midrule
\textbf{\method (ours)}              & \textbf{0.086} & 98.5 & 0.994 & 0.281 \\

\bottomrule
\end{tabular}%
}

\caption{\added{\textbf{Design ablations of \method on the multi-robot formation-control task.} All learned variants share the same intervention point, reward, and environment-step budget; the SAE-reconstruction row involves no steering policy and no training.}}
\label{tab:ablation_formation}
\end{table}

\subsection{Design Ablations}
\label{sec:ablations}
\added{
We isolate three design choices on the formation-control task:
activation editing vs direct policy weight optimization, the PPO action
head for affine editing, and restricting edits to a compact set of
behavior-relevant SAE latents. All learned methods use the same
formation reward and environment-step budget, while all
activation-editing variants intervene at the same layer.
Results appear in Table~\ref{tab:ablation_formation}; learning curves for
the first two ablations are shown in Figure~\ref{fig:ablation}.
}
\added{

\paragraph{Activation editing mitigates catastrophic forgetting.}
We first compare \method with directly optimizing policy weights through
fine-tuning the pretrained base policy or training a new policy from
scratch on the formation reward. Fine-tuning begins competitively,
while the pretrained navigation behavior is still largely retained, but
its performance deteriorates as optimization proceeds
(Fig.~\ref{fig:ablation}a). At convergence, its formation error is
$0.199$, more than twice that of \method, while collision-free rate
drops to $93.5\%$ and success rate to $0.890$. Training from scratch
fails to acquire stable navigation within the same environment-step
budget. In contrast, \method keeps the base policy weights frozen and
improves the formation objective by $\sim3.2\times$ over the base while
largely retaining its navigation competence.

\paragraph{Beta-PPO stabilizes bounded affine editing.}
\method uses a Beta-parameterized PPO action head to produce bounded
affine edits, with multiplicative gains in $[0.5,10.0]$ and additive
offsets in $[-0.1,0.1]$. We compare it against an unbounded Gaussian
head and a Gaussian-tanh head, which enforces the same bounds through
a nonlinear $\tanh$ squashing function. Both alternatives lead to
unstable closed-loop learning (Fig.~\ref{fig:ablation}b):
Gaussian PPO drops to $2.6\%$ collision-free, while Gaussian-tanh
fails completely (Table~\ref{tab:ablation_formation}). The failure of the bounded Gaussian--tanh variant shows that constraining
the edit range alone is insufficient. The Beta head is naturally bounded
with a flexible shape over the interior of the bounds, providing a
well-suited parameterization for the small affine edits used by \method.
This is consistent with the closed-loop behavior in
Sec.~\ref{sec:main_results}: small per-step edits can accumulate through
the closed-loop dynamics to produce substantial behavior-level changes.

% Gaussian PPO drops to $2.6\%$ collision-free, and Gaussian-tanh is catastrophic.
% The editable bounds are tight, and large or saturating edits push the
% activation outside the regime the frozen downstream layers handle
% reliably. The Beta head is naturally bounded with a flexible shape on the
% interior of the bounds, which lets the steering policy explore the
% small-edit regime \method relies on, consistent with the closed-loop
% argument in Sec.~\ref{sec:main_results} that small per-step edits
% accumulated through the closed-loop dynamics produce large behavior
% changes while keeping each step within the activation regime the frozen
% policy was trained on.

\paragraph{SAE and behavior probing improve editing performance.}

We first verify that the SAE itself is behavior-neutral by removing the
steering policy altogether and setting $m_{i,t}=\mathbf{1}$ and
$c_{i,t}=\mathbf{0}$, so the intercepted activation is replaced by its
SAE reconstruction with no steering
(Table~\ref{tab:ablation_formation}). Formation error, collision-free rate, and success remain at base-policy levels ($0.284$, $99.1\%$, $0.990$), confirming that the gains reported for \method come from the learned edits rather than from inserting the SAE.
We then remove the SAE and behavior-probing stage and allow the
steering policy to directly edit every dimension of the intercepted
activation. This increases the steering action dimension from
$2|S_b|$ to $2d$ and removes the restriction to behavior-relevant
features. Even though the intercepted activation is only
$d=30$ dimensional, full-activation editing increases formation error
to $0.797$, reduces the collision-free rate to $66.6\%$, and lowers
success to $0.127$. These results show the benefit of restricting edits
to a compact, probe-selected subset of behavior-relevant SAE latents
rather than directly controlling the full activation.

}

% \begin{table}[t]
% \centering
% \setlength{\tabcolsep}{4pt}
% \renewcommand{\arraystretch}{1.1}
% \resizebox{\columnwidth}{!}{%
% \begin{tabular}{l c c c c}
% \toprule
% \textbf{Method}
%   & \textbf{Form.\ Err.}\,$\downarrow$
%   & CF\,(\%)\,$\uparrow$
%   & SR\,$\uparrow$
%   & Goal\,(m)\,$\downarrow$ \\
% \midrule
% Base Policy                          & 0.280 & 99.5 & 0.997 & 0.271 \\
% \added{SAE reconstruction (no steering)}     & 0.284 & 99.1 & 0.990 & 0.275 \\
% \textbf{\method (ours)}              & \textbf{0.086} & 98.5 & 0.994 & 0.281 \\
% \addlinespace[4pt]
% Fine-tune base policy                & 0.199 & 93.5 & 0.890 & 0.199 \\
% Train from scratch                   & 4.715 & 0.0  & 0.000 & --- \\
% \addlinespace[4pt]
% \method w/ Gaussian PPO              & 0.761 & 2.6  & 0.000 & 0.761 \\
% \method w/ Gaussian-tanh PPO         & 3.276 & 0.0  & 0.000 & --- \\
% \addlinespace[4pt]
% \method w/o SAE (full-activation)    & 0.797 & 66.6 & 0.127 & 0.797 \\
% \bottomrule
% \end{tabular}%
% }

% \caption{\added{\textbf{Design ablations of \method on the multi-robot formation-control task.} All learned variants share the same intervention point, reward, and environment-step budget; the SAE-reconstruction row involves no steering policy and no training.}}
% \label{tab:ablation_formation}
% \{-1.5em}
% \end{table}

\subsection{Real-World Deployment}
\label{sec:realworld}
% \vspace{-0.5em}

We deploy \method on Crazyflie~2.1 quadrotors, a severely resource-constrained platform. We focus on the formation control task and deploy the frozen base policy, SAE, and steering policy zero-shot from simulation. The entire \method pipeline is implemented in C for real-time execution on the STM32 microcontroller while remaining compatible with the 1\,kHz stabilization loop. Each quadrotor performs onboard localization through optical flow,
broadcasts its state to neighbors over a radio link, and generates a local SDF ($2$\,m range) onboard from \emph{a priori}
obstacle positions. All computation runs entirely onboard at $100$\,Hz. Figure~\ref{fig:clae_overview}(b,c) shows representative trajectories: the base policy reaches the goals by splitting the formation between the middle obstacle, while \method keeps the team flying close together as a group retaining the base navigation behavior. 
%==============================================================================
% \vspace{-3pt}
% \section{Conclusion}
% \label{sec:conclusion}
% % \vspace{-5pt}

% We introduced \method, a framework that treats post-training behavior steering of frozen robot policies as a closed-loop problem over SAE latents of an intermediate activation. By learning state-dependent affine edits on a small set of behavior-relevant SAE latents, \method steers individual robot behavior, coordinates multirobot behavior, and supports novel behavior injection, all without touching the base policy weights or action head. Because the interface requires only white-box access to an intermediate activation, we see \method as a step toward a general recipe for adapting pretrained robot policies to deployment-time behavior objectives that were not the primary objective of, or were absent from, the base policy's training, complementing well-established finetuning approaches.

\section{Conclusion}
\label{sec:conclusion}
We introduced \method, a framework that treats post-training behavior steering of frozen robot policies as a closed-loop problem over SAE latents of an intermediate activation. By learning state-dependent affine edits on a small set of behavior-relevant SAE latents, \method steers individual robot behavior, coordinates multirobot behavior, and supports novel behavior injection, all without touching the base policy weights or action head. Because the interface requires only white-box access to an intermediate activation, we see \method as a step toward a general recipe for adapting pretrained robot policies to deployment-time behavior objectives that were not the primary objective of, or were absent from, the base policy's training, complementing well-established finetuning approaches. \added{Two limitations point to future work. \method\ relies on rollout-derived metrics to identify behavior-relevant activations, and for behaviors exogenous to the base policy's objective we used trajectory-modulation proxies; principled alternatives for selecting latents in this regime remain open. We also provide no formal safety or stability guarantees on the edited closed-loop system, and extending the interface to large pretrained models such as VLAs is a natural next step we have not yet explored.} 

\vspace{-0.3em}

\bibliographystyle{IEEEtran}
\bibliography{citations.bib}

@misc{silver2018residual,
  title={Residual Policy Learning},
  author={Silver, Tom and Allen, Kelsey and Tenenbaum, Josh and Kaelbling, Leslie},
  year={2018},
  eprint={1812.06298},
  archivePrefix={arXiv},
  primaryClass={cs.RO}
}

@misc{swann2026sparse,
  title={Sparse Autoencoders Reveal Interpretable and Steerable Features in VLA Models},
  author={Swann, Aiden and McGranahan, Lachlain and Buurmeijer, Hugo and Kennedy, Monroe and Schwager, Mac},
  year={2026},
  eprint={2603.19183},
  archivePrefix={arXiv},
  primaryClass={cs.RO}
}

@misc{alain2016understanding,
  title         = {Understanding Intermediate Layers Using Linear Classifier Probes},
  author        = {Alain, Guillaume and Bengio, Yoshua},
  year          = {2016},
  eprint        = {1610.01644},
  archivePrefix = {arXiv},
  primaryClass  = {stat.ML}
}

@inproceedings{kim2018tcav,
  title     = {Interpretability Beyond Feature Attribution: Quantitative Testing with Concept Activation Vectors ({TCAV})},
  author    = {Kim, Been and Wattenberg, Martin and Gilmer, Justin and Cai, Carrie and Wexler, James and Vi{\'e}gas, Fernanda and Sayres, Rory},
  booktitle = {Proceedings of the 35th International Conference on Machine Learning},
  pages     = {2668--2677},
  year      = {2018},
  volume    = {80},
  series    = {Proceedings of Machine Learning Research},
  publisher = {PMLR}
}

@article{belinkov2022probing,
  title   = {Probing Classifiers: Promises, Shortcomings, and Advances},
  author  = {Belinkov, Yonatan},
  journal = {Computational Linguistics},
  volume  = {48},
  number  = {1},
  pages   = {207--219},
  year    = {2022},
  doi     = {10.1162/coli_a_00422}
}

@misc{cunningham2023sparse,
  title         = {Sparse Autoencoders Find Highly Interpretable Features in Language Models},
  author        = {Cunningham, Hoagy and Ewart, Aidan and Riggs, Logan and Huben, Robert and Sharkey, Lee},
  year          = {2023},
  eprint        = {2309.08600},
  archivePrefix = {arXiv},
  primaryClass  = {cs.LG}
}

@misc{turner2023actadd,
  title={Steering Language Models With Activation Engineering},
  author={Turner, Alexander Matt and Thiergart, Lisa and Leech, Gavin and Udell, David and Vazquez, Juan J. and Mini, Ulisse and MacDiarmid, Monte},
  year={2023},
  eprint={2308.10248},
  archivePrefix={arXiv},
  primaryClass={cs.CL}
}

@misc{zou2023representation,
  title={Representation Engineering: A Top-Down Approach to AI Transparency},
  author={Zou, Andy and Phan, Long and Chen, Sarah and Campbell, James and Guo, Phillip and Ren, Richard and Pan, Alexander and Yin, Xuwang and Mazeika, Mantas and others},
  year={2023},
  eprint={2310.01405},
  archivePrefix={arXiv},
  primaryClass={cs.LG}
}

@misc{singh2024representation_surgery,
  title={Representation Surgery: Theory and Practice of Affine Steering},
  author={Singh, Shashwat and Ravfogel, Shauli and Herzig, Jonathan and Aharoni, Roee and Cotterell, Ryan and Kumaraguru, Ponnurangam},
  year={2024},
  eprint={2402.09631},
  archivePrefix={arXiv},
  primaryClass={cs.LG}
}

@article{liu2026vls,
  title     = {VLS: Steering Pretrained Robot Policies via Vision-Language Models},
  author    = {Shuo Liu and Ishneet Sukhvinder Singh and Yiqing Xu and Jiafei Duan and Ranjay Krishna},
  journal   = {arXiv preprint arXiv:2602.03973},
  year      = {2026}
}

@article{wu2025forewarn,
  title={From foresight to forethought: Vlm-in-the-loop policy steering via latent alignment},
  author={Wu, Yilin and Tian, Ran and Swamy, Gokul and Bajcsy, Andrea},
  journal={Robotics: Science and Systems (RSS)},
  year={2025}

}

@misc{chen2026steerable,
  title={Steerable Vision-Language-Action Policies for Embodied Reasoning and Hierarchical Control},
  author={Chen, William and Bhatia, Jagdeep Singh and Glossop, Catherine and Mathihalli, Nikhil and Doshi, Ria and Tang, Andy and Driess, Danny and Pertsch, Karl and Levine, Sergey},
  year={2026},
  eprint={2602.13193},
  archivePrefix={arXiv},
  primaryClass={cs.RO}
}

@misc{haon2025mechanistic,
  title={Mechanistic Interpretability for Steering Vision-Language-Action Models},
  author={H{\"a}on, Bear and Stocking, Kaylene and Chuang, Ian and Tomlin, Claire},
  year={2025},
  eprint={2509.00328},
  archivePrefix={arXiv},
  primaryClass={cs.RO}
}

@misc{das2025latent,
  title={Latent Activation Editing: Inference-Time Refinement of Learned Policies for Safer Multirobot Navigation},
  author={Das, Satyajeet and Chiu, Darren and Huang, Zhehui and Lindemann, Lars and Sukhatme, Gaurav S.},
  year={2025},
  eprint={2509.20623},
  archivePrefix={arXiv},
  primaryClass={cs.RO}
}

@misc{johannink2018residual,
  title={Residual Reinforcement Learning for Robot Control},
  author={Johannink, Tobias and Bahl, Shikhar and Nair, Ashvin and Luo, Jianlan and Kumar, Avinash and Loskyll, Matthias and Ojea, Juan Aparicio and Solowjow, Eugen and Levine, Sergey},
  year={2018},
  eprint={1812.03201},
  archivePrefix={arXiv},
  primaryClass={cs.RO}
}

@article{nakamoto2024vgps,
  author    = {Mitsuhiko Nakamoto and Oier Mees and Aviral Kumar and Sergey Levine},
  title     = {Steering Your Generalists: Improving Robotic Foundation Models via Value Guidance},
  journal   = {Conference on Robot Learning (CoRL)},
  year      = {2024},
}

@inproceedings{wang2024itps,
  title={Inference-time policy steering through human interactions},
  author={Wang, Yanwei and others},
  booktitle={2025 IEEE ICRA},
}

@misc{wagenmaker2025dsrl,
  title={Steering Your Diffusion Policy with Latent Space Reinforcement Learning},
  author={Wagenmaker, Andrew and Zhang, Yunchu and Nakamoto, Mitsuhiko and Park, Seohong and Yagoub, Waleed and Nagabandi, Anusha and Gupta, Abhishek and Levine, Sergey},
  year={2025},
  eprint={2506.15799},
  archivePrefix={arXiv},
  primaryClass={cs.RO}
}

@article{bricken2023towards,
  title={Towards Monosemanticity: Decomposing Language Models With Dictionary Learning},
  author={Bricken, Trenton and others},
  journal={Transformer Circuits Thread},
  year={2023},
}

@article{buurmeijer2026observing,
  title={Observing and Controlling Features in Vision-Language-Action Models},
  author={Buurmeijer, Hugo and Alonso, Carmen Amo and Swann, Aiden and Pavone, Marco},
  journal={arXiv preprint arXiv:2603.05487},
  year={2026}
}

@article{huang2023quadswarm,
  title={Quadswarm: A modular multi-quadrotor simulator for deep reinforcement learning with direct thrust control},
  author={Huang, Zhehui and Batra, Sumeet and Chen, Tao and Krupani, Rahul and Kumar, Tushar and Molchanov, Artem and Petrenko, Aleksei and Preiss, James A and Yang, Zhaojing and Sukhatme, Gaurav S},
  journal={arXiv preprint arXiv:2306.09537},
  year={2023}
}

@inproceedings{huang2024collision,
  title={Collision avoidance and navigation for a quadrotor swarm using end-to-end deep reinforcement learning},
  author={Huang, Zhehui and Yang, Zhaojing and Krupani, Rahul and {\c{S}}enba{\c{s}}lar, Bask{\i}n and Batra, Sumeet and Sukhatme, Gaurav S},
  booktitle={IEEE Int. Conf. Robot. Autom. (ICRA)},
  year={2024},
}

@inproceedings{petrazzini2021proximal,
  title={Proximal policy optimization with continuous bounded action space via the beta distribution},
  author={Petrazzini, Irving GB and Antonelo, Eric A},
  booktitle={symposium series on computational intelligence (SSCI)},
  year={2021},
  organization={IEEE}
}

@inproceedings{mellinger2011minimum,
  title={Minimum snap trajectory generation and control for quadrotors},
  author={Mellinger, Daniel and Kumar, Vijay},
  booktitle={2011 IEEE international conference on robotics and automation},
  pages={2520--2525},
  year={2011},
  organization={Ieee}
}

@inproceedings{wang2016safety,
  title={Safety barrier certificates for heterogeneous multi-robot systems},
  author={Wang, Li and Ames, Aaron and Egerstedt, Magnus},
  booktitle={Amer. cont. conf. (ACC)},
  year={2016},
}

@book{templeton2024scaling,
  title={Scaling monosemanticity: Extracting interpretable features from claude 3 sonnet},
  author={Templeton, Adly},
  year={2024},
  publisher={Anthropic}
}

@inproceedings{ross2011reduction,
  title     = {A Reduction of Imitation Learning and Structured Prediction to
               No-Regret Online Learning},
  author    = {Ross, St{\'e}phane and Gordon, Geoffrey J. and Bagnell, J. Andrew},
  booktitle = {Proc. Int. Conf. Artif. Intell. Stat. (AISTATS)},
 year = {2011}, publisher = {PMLR}
}

@inproceedings{chi2023diffusion,
  title     = {Diffusion Policy: Visuomotor Policy Learning via Action Diffusion},
  author    = {Chi, Cheng and Feng, Siyuan and Du, Yilun and Xu, Zhenjia and
               Cousineau, Eric and Burchfiel, Benjamin and Song, Shuran},
  booktitle = {Proceedings of Robotics: Science and Systems (RSS)},
  year      = {2023}, doi = {10.15607/RSS.2023.XIX.026}
}

@article{loquercio2021agile,
  title   = {Learning high-speed flight in the wild},
  author  = {Loquercio, Antonio and Kaufmann, Elia and Ranftl, Ren{\'e} and
             M{\"u}ller, Matthias and Koltun, Vladlen and Scaramuzza, Davide},
  journal = {Science Robotics}, volume = {6}, number = {59},
 year = {2021}
}

@inproceedings{czarnecki2019distilling,
  title={Distilling policy distillation},
  author={Czarnecki, Wojciech M and Pascanu, Razvan and Osindero, Simon and Jayakumar, Siddhant and Swirszcz, Grzegorz and Jaderberg, Max},
  booktitle={international conference on artificial intelligence and statistics},
  year={2019},
  organization={PMLR}
}
\clearpage

\appendix

\subsection{Steering Policy Details}
\label{app:steering_details}

This subsection specifies the steering-policy observations and action
constraints used in our experiments. During steering-policy training,
the base policy $\pi_{\theta_0}$ and SAE encoder--decoder
$(E_\psi,D_\psi)$ remain fixed; only the steering policy is optimized.

\subsubsection{Steering Policy Action}
\label{app:steering_action}

For target behavior $b$, let $\mathcal{S}_b$ denote the selected
behavior-relevant SAE latent set. At each timestep, the steering policy
outputs
\begin{equation}
    \xi_{i,t}
    =
    (m_{i,t},c_{i,t})
    \in
    \mathbb{R}^{2|\mathcal{S}_b|},
\end{equation}
where $m_{i,t}$ and $c_{i,t}$ are the multiplicative and additive
components of the affine edit in Eq.~\eqref{eq:affine_edit}.
Only latents in $\mathcal{S}_b$ are modified; all remaining SAE latents
are left unchanged before decoding through the frozen SAE decoder.
The steering policy therefore influences the base controller only
through its internal activation representation and does not directly
output motor commands or action residuals.

For the multiplicative-only ablation, we set
$c_{i,t}=\mathbf{0}$ and learn only
$m_{i,t}\in\mathbb{R}^{|\mathcal{S}_b|}$.

\paragraph{Bounded PPO action head.}
We use a Beta-parameterized continuous action head for the steering
policy. During rollout, multiplicative gains are clipped to
$[0.5,10.0]$ and additive offsets to $[-0.1,0.1]$.
These bounds are implementation hyperparameters used to limit the
magnitude of activation interventions; they are not constraints imposed
by the \method formulation and do not provide a safety or stability
guarantee.

\subsubsection{Steering Observations}
\label{app:steering_observations}

Each steering policy combines a shared latent-control context with a
compact task-specific observation head. The shared context contains the
current values of the selected SAE latents, the previous affine edit, a
history of the selected latents, and auxiliary features derived from the
SAE and frozen-policy representations.
This shared component is held
fixed across tasks; Table~\ref{tab:app_obs_heads} lists the
task-specific features that differ across behaviors.

\begin{table}[h]
\centering
\caption{Task-specific steering observation heads. Each head is
concatenated with the shared latent-control context.}
\label{tab:app_obs_heads}
\begin{tabular}{ll}
\toprule
Task & Task-specific observation head \\
\midrule
Velocity &
$\left[v_{i,t}-v^\star_{i,t},\; p_{i,t},\; g_{i,t}\right]$ \\
Formation &
$\left[v_{i,t},\; g_{i,t}-p_{i,t},\; p_{i,t},\;
\phi_{i,1},\phi_{i,2},\phi_{i,3},\; s^\star\right]$ \\
Stealth &
$\left[p_{i,t},\; v_{i,t},\; g_{i,t}-p_{i,t},\;
\mathbf{d}^{\mathrm{cam}}_{i,t}\right]$ \\
\bottomrule
\end{tabular}
\end{table}

Here $p_{i,t}$, $v_{i,t}$, and $g_{i,t}$ denote the robot position,
velocity, and goal, respectively, and $v^\star_{i,t}$ is the reference
velocity for velocity tracking. For formation control, $s^\star$ is the
target square side length and $\phi_{i,r}$ describes the $r$-th nearest
teammate within the same four-robot group:
\begin{equation}
    \phi_{i,r}
    =
    \left[
    d_{i,r},
    p^{xy}_{j_r,t}-p^{xy}_{i,t},
    v^{xy}_{j_r,t}-v^{xy}_{i,t}
    \right],
\end{equation}
where $j_r$ denotes the corresponding teammate and $d_{i,r}$ is its
XY distance from robot $i$.

For stealth navigation,
$\mathbf{d}^{\mathrm{cam}}_{i,t}\in\mathbb{R}^{9}$ is a local
camera-distance field evaluated on a fixed $3\times3$ neighborhood
around the robot's XY position. Each entry contains an SDF-like signed
distance to the nearest surveillance-region boundary represented in
the environment encoding, with negative values inside the represented
region, zero on its boundary, and positive values outside. This provides
the steering policy with a compact local representation of surveillance
geometry without directly supplying the complete camera map.

All components of the steering observation are available at deployment
time from the robot state estimate, task specification, neighboring-agent
information, environment context, or the frozen policy's activation
stream; no future trajectory information is provided to the steering
policy.

\subsection{Task-Reward Instantiations}
\label{app:task_rewards}

Section~\ref{sec:learning_steering_policy} defines the common steering
objective. Here we specify its behavior-dependent reward terms.
% Table~\ref{tab:reward_hparams} lists the reported reward coefficients,
% which are experimental hyperparameters selected during preliminary tuning.
Tunable reward coefficients are kept symbolic in the equations and
listed in Table~\ref{tab:reward_hparams}. These values were selected during
preliminary tuning for stable PPO learning and validation performance; they are
experimental hyperparameters rather than constraints imposed by the CLAE
formulation.

Across all tasks, the per-drone steering reward follows
\begin{equation}
    r^{(b)}_{i,t}
    =
    r^{(b)}_{\mathrm{task},i,t}
    +
    w^{(b)}_{\mathrm{stab}} r^{\mathrm{stab}}_{i,t}
    -
    \lambda^{(b)}_{\mathrm{crash}} C_{i,t},
    \label{eq:app_common_reward}
\end{equation}

\paragraph{Velocity steering.}
The velocity task steers each drone toward a reference velocity
$v^\star_{i,t}$. The task reward is the negative tracking error,
\begin{equation}
    r^{\mathrm{track}}_{i,t}
    =
    -
    \left\|
    v_{i,t} - v^\star_{i,t}
    \right\|_1,
    \qquad
    r^{\mathrm{vel}}_{\mathrm{task},i,t}
    =
    w_{\mathrm{track}}
    r^{\mathrm{track}}_{i,t}.
    \label{eq:app_velocity_reward}
\end{equation}

\paragraph{Formation steering.}
The formation task steers each four-drone group toward square geometry while
preserving navigation progress. For a group $G$, let
$d_{(1)} \leq \cdots \leq d_{(6)}$ be the sorted pairwise XY distances. The
group-level square error is
\begin{equation}
    E_G
    =
    \frac{1}{6}
    \left(
    \sum_{a=1}^{4}
    \frac{|d_{(a)}-s^\star|}{s^\star}
    +
    \sum_{a=5}^{6}
    \frac{|d_{(a)}-\sqrt{2}s^\star|}
    {\sqrt{2}s^\star}
    \right),
    \label{eq:app_group_square_error}
\end{equation}
where $s^\star$ is the target side length.

For drone $i$, let $r_{i,(1)}, r_{i,(2)}, r_{i,(3)}$ be its sorted XY distances
to the other three drones in the same group. The local neighbor error is
\begin{equation}
    E^{\mathrm{nbr}}_i
    =
    \sum_{a=1}^{2}
    \frac{|r_{i,(a)}-s^\star|}{s^\star}
    +
    \frac{|r_{i,(3)}-\sqrt{2}s^\star|}
    {\sqrt{2}s^\star}.
    \label{eq:app_neighbor_error}
\end{equation}
Let $\bar{p}_{G,t}$ be the group center. The radial error is
\begin{equation}
    E^{\mathrm{rad}}_i
    =
    \frac{
    \left|
    \|p^{xy}_{i,t}-\bar{p}^{xy}_{G,t}\|_2
    -
    s^\star/\sqrt{2}
    \right|
    }
    {s^\star/\sqrt{2}}.
    \label{eq:app_radial_error}
\end{equation}
The local formation error and shape reward are
% \begin{equation}
%     E^{\mathrm{local}}_i
%     =
%     \beta_{\mathrm{nbr}} E^{\mathrm{nbr}}_i
%     +
%     \beta_{\mathrm{rad}} E^{\mathrm{rad}}_i,
%     \qquad
%     r^{\mathrm{shape}}_{i,t}
%     =
%     -
%     \left(
%     \alpha_{\mathrm{grp}} E_G
%     +
%     \alpha_{\mathrm{loc}} E^{\mathrm{local}}_i
%     \right).
%     \label{eq:app_shape_reward}
% \end{equation}

\begin{equation}
\begin{aligned}
    E^{\mathrm{local}}_i
    &=
    \beta_{\mathrm{nbr}} E^{\mathrm{nbr}}_i
    +
    \beta_{\mathrm{rad}} E^{\mathrm{rad}}_i, \\
    r^{\mathrm{shape}}_{i,t}
    &=
    -
    \left(
    \alpha_{\mathrm{grp}} E_G
    +
    \alpha_{\mathrm{loc}} E^{\mathrm{local}}_i
    \right).
\end{aligned}
\label{eq:app_shape_reward}
\end{equation}
The goal-distance term penalizes normalized distance to the assigned goal:
\begin{equation}
    r^{\mathrm{goal}}_{i,t}
    =
    -
    \frac{
    \|g_{i,t}-p_{i,t}\|_2
    }
    {
    \|g_{i,0}-p_{i,0}\|_2
    }.
    \label{eq:app_goal_reward}
\end{equation}
The formation task reward is
\begin{equation}
    r^{\mathrm{form}}_{\mathrm{task},i,t}
    =
    w_{\mathrm{shape}}
    r^{\mathrm{shape}}_{i,t}
    +
    w_{\mathrm{goal}}
    r^{\mathrm{goal}}_{i,t}.
    \label{eq:app_formation_task_reward}
\end{equation}

\paragraph{Stealth steering.}
The stealth navigation task introduces camera surveillance regions that were
not part of the base-policy training objective. The task reward combines a sparse exposure penalty, a dense margin penalty, and a smaller goal-directed reward.
The exposure term penalizes entry into a valid surveillance footprint,
while the margin term provides a smoother learning signal near camera-facing
regions.
Let $c_m$ denote the center of camera region $m$, $R$ its surveillance radius,
and $H_m$ the valid camera-facing half-plane. A drone is exposed to camera $m$
when it lies inside both the surveillance radius and the valid camera-facing
region:
\begin{equation}
    I_{i,m,t}
    =
    \mathbf{1}
    \left[
    \|p^{xy}_{i,t}-c_m\|_2 \le R
    \;\wedge\;
    p^{xy}_{i,t}\in H_m
    \right].
    \label{eq:app_camera_indicator}
\end{equation}
The zone penalty is
\begin{equation}
    r^{\mathrm{zone}}_{i,t}
    =
    -
    \sum_m I_{i,m,t}.
    \label{eq:app_zone_penalty}
\end{equation}

Because the binary exposure penalty is sparse, we also use a soft margin around
camera-facing regions. Let $R_{\mathrm{out}}=R+\delta$ and define
\begin{equation}
    \eta_{i,m,t}
    =
    \mathrm{clip}
    \left(
    \frac{
    R_{\mathrm{out}}
    -
    \|p^{xy}_{i,t}-c_m\|_2
    }
    {
    R_{\mathrm{out}}
    },
    0,
    1
    \right).
    \label{eq:app_camera_margin}
\end{equation}
The value $\eta_{i,m,t}$ is zero outside the outer margin and increases as the
drone approaches the camera center. The margin penalty is
\begin{equation}
    r^{\mathrm{margin}}_{i,t}
    =
    -
    \sum_m
    \mathbf{1}
    \left[
    p^{xy}_{i,t}\in H_m,\;
    \|p^{xy}_{i,t}-c_m\|_2 \le R_{\mathrm{out}}
    \right]
    \eta_{i,m,t}^{\kappa}.
    \label{eq:app_camera_margin_penalty}
\end{equation}
Thus, within the camera-facing half-plane, the margin term provides a dense
signal before and inside the surveillance radius: states farther from the
camera center receive little or no penalty, whereas states closer to the
camera center receive a larger penalty. Entering a
surveillance region affects reward but does not terminate the episode.
The stealth task reward is
\begin{equation}
    r^{\mathrm{stealth}}_{\mathrm{task},i,t}
    =
    w_{\mathrm{zone}}
    r^{\mathrm{zone}}_{i,t}
    +
    w_{\mathrm{margin}}
    r^{\mathrm{margin}}_{i,t}
    +
    w_{\mathrm{goal}}^{\mathrm{stealth}}
    r^{\mathrm{goal}}_{i,t}.
    \label{eq:app_stealth_task_reward}
\end{equation}
% The stealth task reward is
% \begin{equation}
%     r^{\mathrm{stealth}}_{\mathrm{task},i,t}
%     =
%     w_{\mathrm{zone}}
%     r^{\mathrm{zone}}_{i,t}
%     +
%     w_{\mathrm{margin}}
%     r^{\mathrm{margin}}_{i,t}.
%     \label{eq:app_stealth_task_reward}
% \end{equation}
The goal-distance term uses a small weight and serves only to bias the edited
policy toward reaching the goal; the exposure terms remain the primary stealth
objective. The reward coefficients are kept symbolic in the definitions above. Table~\ref{tab:reward_hparams}
lists the values used in the reported PPO steering experiments.  In the reported experiments, each
camera surveillance region has a diameter of $5\,\mathrm{m}$.
% These values were
% selected during preliminary tuning and are not part of the CLAE formulation.

% \begin{table}[h]
% \centering
% \caption{Reward hyperparameters used in the reported PPO steering experiments.}
% \label{tab:reward_hparams}
% \begin{tabular}{@{}lc@{}}
% \toprule
% Hyperparameter & Value \\
% \midrule
% $w_{\mathrm{track}}$ & $2$ \\
% $w_{\mathrm{shape}}$ & $1$ \\
% $w_{\mathrm{goal}}$ & $1$ \\
% $w_{\mathrm{zone}}$ & $0.5$ \\
% $w_{\mathrm{margin}}$ & $1$ \\
% $w_{\mathrm{stab}}$ & $2$ \\
% $\lambda_{\mathrm{crash}}$ & $50$ \\
% $\alpha_{\mathrm{grp}},\,\alpha_{\mathrm{loc}}$ & $0.6,\,0.4$ \\
% $\beta_{\mathrm{nbr}},\,\beta_{\mathrm{rad}}$ & $0.75,\,0.25$ \\
% \bottomrule
% \end{tabular}
% \end{table}
\begin{table}[h]
\centering
\begin{tabular}{@{}lc@{}}
\toprule
Coefficient & Value \\
\midrule
$w_{\mathrm{track}}$ & $2$ \\
$w_{\mathrm{shape}}$ & $1$ \\
$w_{\mathrm{goal}}$ & $1$ \\
$w_{\mathrm{zone}}$ & $0.5$ \\
$w_{\mathrm{margin}}$ & $1$ \\
$w_{\mathrm{stab}}$ & $2$ \\
$\lambda_{\mathrm{crash}}$ & $50$ \\
\addlinespace[2pt]
$\alpha_{\mathrm{grp}},\,\alpha_{\mathrm{loc}}$ & $0.6,\,0.4$ \\
$\beta_{\mathrm{nbr}},\,\beta_{\mathrm{rad}}$ & $0.75,\,0.25$ \\
\bottomrule
\end{tabular}
\caption{Reward coefficients used in the reported PPO steering experiments.}
\label{tab:reward_hparams}
\end{table}
\subsection{Sparse Latent Probe Metrics}
\label{app:latent_selection}

Section~\ref{sec:sparse_basis}
describes the split-stable probing procedure used to select
behavior-relevant SAE latents. Here we summarize the rollout metrics used as
post-hoc probes for each target behavior. These metrics are computed only after SAE training and do not enter the SAE reconstruction-and-sparsity objective.

For velocity steering, probes use the velocity components and scalar speed. For
formation steering, probes use relational square-geometry quality and
inter-drone spacing within each four-drone group. For stealth steering, where
camera avoidance is not part of the frozen policy's original training objective,
probes use trajectory-modulation metrics, including lateral
motion, heading change, goal progress, clearance, and collision risk. Selected latent indices are SAE-basis-specific; we therefore report the post-hoc selection procedure and rollout metrics used for ranking rather than checkpoint-specific indices.

\subsection{Baseline and CLAE Variant Details}
\label{sec:baseline_details}
This subsection supplements the comparisons in
Section~\ref{sec:methods} with implementation details for the baselines
and single-component ablations.

\subsubsection{Prior activation-editing baselines}
Fixed Scalar SAE Steering~\cite{swann2026sparse} applies a constant
activation shift along the decoder direction of the top
behavior-associated SAE latent.
Linear Probe Intervention~\cite{buurmeijer2026observing} fits an
offline linear probe and applies a minimum-norm perturbation toward
the target probe value. Neither method learns a steering policy
from closed-loop return.

\subsubsection{Single-component ablations}
Closed-Loop Multiplicative Editing fixes $c_t=\mathbf{0}$ and learns
the gains $m_t$, whereas Closed-Loop Additive-Only Editing fixes
$m_t=\mathbf{1}$ and learns the offsets $c_t$.
Both retain state-dependent closed-loop editing at the same
activation-intervention point as \method.
Unselected SAE latents remain unchanged.

\subsubsection{Residual Action RL}
Residual Action RL~\cite{johannink2018residual} learns a bounded
additive correction to the frozen base policy's normalized thrust
command. As described in Section~\ref{sec:methods}, it uses the same
steering observations, task reward, PPO algorithm, and environment-step
budget as \method.

\subsubsection{Distilled Diffusion Policy}
\label{app:diffusion_policy}

We train a decentralized diffusion policy~\cite{chi2023diffusion} by
behavior cloning demonstrations generated by the frozen base policy.
The training corpus contains 10,000 eight-robot episodes, each lasting
1,500 simulator steps, corresponding to 80,000 individual robot
trajectories. 

The diffusion policy shares parameters across robots and
conditions on a short history of the same per-agent observations used by
the base RL policy. A conditional temporal U-Net predicts a sequence of 16
four-dimensional motor commands. At inference, we use the
exponential-moving-average model with ten deterministic DDIM denoising
steps. The first four generated actions are executed sequentially before
replanning. The unsteered diffusion policy and DSRL use the same
diffusion-model inference protocol.

\subsubsection{DSRL Noise Steering}
\label{app:dsrl}

We implement a DSRL-NA-style noise-steering baseline following
\cite{wagenmaker2025dsrl}. A separate steering policy is trained for each
behavior while the distilled diffusion policy remains frozen.
The steering policy conditions on the diffusion policy's observation
history together with behavior-specific task features and predicts the
bounded initial noise supplied to the frozen denoiser. DSRL therefore
steers the generated action sequence through diffusion noise rather than
directly modifying the final motor commands.

We use the same behavior-specific rewards as \method for velocity
tracking, formation control, and stealth navigation.
Our implementation uses replay-paired action-to-noise value distillation:
action-value critics learn from the executed action chunks, while
noise-value critics learn the corresponding values in diffusion-noise
space. The steering actor is subsequently optimized against the
noise-value critics using an entropy-regularized objective.
The diffusion policy remains fixed throughout DSRL training and
evaluation.
All DSRL policies are trained and evaluated with the same frozen
diffusion-policy checkpoint used for the unsteered diffusion baseline.

\end{document}